\documentclass[twoside]{article}

\usepackage[accepted]{aistats2023Modified}
\usepackage[utf8]{inputenc} 
\usepackage[T1]{fontenc}    
\usepackage{hyperref}       
\usepackage{url}            
\usepackage{booktabs}       
\usepackage{amsfonts}       
\usepackage{nicefrac}       
\usepackage{microtype}      
\usepackage{xcolor}         
\usepackage[round]{natbib}

\usepackage{microtype}
\usepackage{subfigure}

\usepackage{graphicx}
\usepackage{amsmath}
\usepackage{amssymb}
\usepackage{mathtools}
\usepackage{algorithm}
\usepackage{algorithmic}
\usepackage{amsfonts}
\usepackage{textcomp}
\usepackage{bm}

\newtheorem{Theorem}{Theorem}[section]
\newtheorem{proof}{Proof}[section]
\newtheorem{Problem}{Problem}[section]
\newtheorem{corollary}{Corollary}[section]
\newtheorem{lemma}{Lemma}[section]

\newtheorem{Assumption}{Assumption}[section]

\newtheorem{Definition}{Definition}[section]
\newtheorem{Remark}{Remark}[section]
\newcommand{\e}{\mathbf{e}}

\newcommand{\x}{\mathbf{x}}
\newcommand{\y}{\mathbf{y}}
\newcommand{\z}{\mathbf{z}}
\newcommand{\w}{\mathbf{w}}

\newcommand{\reals}{\mathbb{R}}

\newcommand{\hyf}{\hat{\y}_{f}}
\newcommand{\hyfi}{\hyf}

\newcommand{\bP}{\mathbf{P}}
\newcommand{\bE}{\mathbf{E}}
\newcommand{\F}{\mathbf{F}}
\newcommand{\bnu}{\bm{\nu}}
\newcommand{\etab}{\bm{\eta}}

\DeclareMathOperator*{\argmin}{\mathrm{argmin}}

\title{PAC-Bayesian-Like Error Bound for a Class of Linear Time-Invariant Stochastic State-Space Models}

\begin{document}
%

%

\twocolumn[

\aistatstitle{PAC-Bayesian-Like Error Bound for a Class of Linear Time-Invariant Stochastic State-Space Models}

\aistatsauthor{Deividas Eringis \And John Leth \And Zheng-Hua Tan \And Rafal Wisniewski } %

\aistatsaddress{ Aalborg University \And  Aalborg University \And Aalborg University \And Aalborg University }

\aistatsauthor{Mihaly Petreczky }
\aistatsaddress{ Laboratoire Signal et Automatique de Lille (CRIStAL) }

] 
\begin{abstract}
	In this paper we derive a PAC-Bayesian-Like error bound for a class of stochastic dynamical systems with inputs, namely, for linear time-invariant stochastic state-space models (stochastic LTI systems for short). This class of systems is  widely used in control engineering and econometrics, in particular, they represent a special case of recurrent neural networks. In this paper we  1) formalize the learning problem for stochastic LTI systems with inputs, 2) derive a PAC-Bayesian-Like error bound for such systems, 3) discuss various consequences of this error bound.
\end{abstract}
\section{Introduction}
 The class of stochastic LTI systems in
state-space form
is 
widely used in control engineering and econometrics  to model time-series,
and it has a rich literature on learning 
\cite{LjungBook}. 
However, there are no results on PAC-Bayesian error bounds for stochastic LTI systems with inputs.\par
The Probably Approximately Correct (PAC)-Bayesian learning theory 
is an important tool for analysing theoretical properties of machine learning algorithm, see  \cite{guedj2019primer,alquier2021userfriendly,zhang-06,grunwald-2012,alquier-15,nips-16,ShethK17}. 
\par
In this paper we will present two PAC-Bayesian-Like error bounds for stochastic LTI systems in state-space form with inputs. One bound is based on Kullback-Leibler (KL)
divergence, the other based on R\'enyi divergence.
The bound involving R\'enyi divergence converges to zero as number of data points $N \rightarrow \infty$ with the rate $O(\frac{1}{\sqrt{N}})$. 
The bound involving KL-divergence converges to a problem dependant constant with the rate $O(\frac{1}{N})$. 
The systems considered are assumed to be in state-space form, to have unbounded (Gaussian) inputs and noises. In addition, we use quadratic loss function, and the prediction error is considered on an infinite time horizon.
\par
\textbf{Motivation}
PAC and PAC-Bayesian bounds have been a major tool for analyzing
learning algorithms. 
They provide bounds on the generalization error in terms of the empirical error,
in a manner which is independent of the learning algorithm. Hence, these bounds can be used to analyze and explain a wide variety of learning algorithms. Moreover,
by minimizing the error bound, new, theoretically well-founded learning algorithms can be 
formulated. In particular, PAC-Bayesian error bounds turned out to be useful for 
providing non-vacuous error bounds for neural networks \cite{Dziugaite2017}.
\par
  While there is a wealth of literature on PAC \cite{shalev2014understanding} and PAC-Bayesian \cite{alquier2021userfriendly,guedj2019primer},
  bounds for static models, much less is known on  dynamical systems.
  \par
  The motivation for the choice of the class of LTI systems is as follows. First,  LTI systems are among the simplest class of dynamical systems in state-space form, and PAC-Bayesian bounds
  for them could help to derive such bounds for
  more general classes of systems, for instance 
  for recurrent neural networks (RNN). 
  Note that LTI systems are a subset of RNNs. 
  Second, PAC-Bayesian-Like bounds could be interesting for learning LTI systems.
  Traditionally, the literature on LTI systems \cite{LjungBook} has focused on statistical consistency. More recently, several results have appeared on finite-sample bounds for learning LTI systems, but they are valid only for specific learning algorithms or for very limited subclasses \cite{simchowitz2021statistical},
  \par
\textbf{Related work}
   The related literature can be divided into the following categories.
  \\
   \textbf{Generalization bounds for RNNs.}
      PAC bounds for RNN wered developed in \cite{KOIRAN199863,sontag1998learning,pmlr-v108-chen20d} using VC dimension, and
      in  \cite{WeiRNN,Akpinar_Kratzwald_Feuerriegel_2020,pmlr-v161-joukovsky21a, pmlr-v108-chen20d} using Rademacher complexity, and in 
      \cite{pmlr-v80-zhang18g} using PAC-Bayesian bounds approach.
      However, all the cited papers assume noiseless models,  
      a fixed number of time-steps, that the training data are i.i.d sampled time-series, and the signals are bounded.
      In contrast, we consider (1) noisy models, (2) prediction error defined on infinite time horizon, (3) only one single time series available for training data, and (4) unbounded signals. 
      Moreover, several
      papers \cite{KOIRAN199863,sontag1998learning,pmlr-v144-hanson21a} assume
      Lipschitz loss functions, while we use quadratic loss function.
     \par
\textbf{PAC and PAC-Bayesian bounds for autoregressive models.} 
PAC bounds for linear dynamical systems in autoregressive form were proposed in \cite{campi2002finite,vidyasagar2006learning}.
The cited papers assumed bounded inputs, \cite{vidyasagar2006learning} bounded loss, and \cite{campi2002finite} assumed bounded inputs and noise and finite horizon prediction. 
Moreover, \cite{vidyasagar2006learning} restricts attention to a small subset of linear systems in input-output form, and the error bound of \cite{campi2002finite} is exponential in the number of parameters. None of the cited papers covers the class of stochastic LTI systems in state-space form. 
In  \cite{alquier2012pred,alquier2013prediction} auto-regressive models without exogenous inputs were considered, and the variables were either assumed to be bounded or the loss function was assumed to be Lipschitz. In contrast, we consider state-space models with inputs, 
the variables are not bounded and the loss function is quadratic. 
That is, the learning problem considered in this paper is
different from that of \cite{alquier2012pred,alquier2013prediction}.
PAC-Bayesian bounds for
autoregressive models with exogenous inputs were developed in \cite{shalaeva2019improved}.
In contrast to \cite{shalaeva2019improved} 
we consider state-space models.
Moreover, the error bound of this paper is tighter.
\par
Note that in contrast to autoregressive models, 
state-space models use an infinite past of the inputs and past observations to generate predictions, see Remark \ref{remark:ARMAdifference}.
This required new approaches in comparison to the cited papers. 
\par
\textbf{PAC-Bayesian bounds for state-space representation.}
In \cite{haussmann2021learning} learning of stochastic differential equations without inputs was considered and it was assumed that  several independently sampled time-series were available for learning. 
In contrast, in this paper we deal with discrete-time systems with inputs and the learning takes place from a single time-series.
In \cite{PACMarkov} learning of general Markov-chains was considered, but the state of the Markov-chain was assumed to be observable and no inputs were considered. The learning problem of \cite{PACMarkov} is thus different from the one considered in this paper.
\par
In \cite{CDC21paper} 
PAC-Bayesian error bounds were developed for autonomous LTI state-space systems without exogenous input. 
In contrast to \cite{CDC21paper}, in the current paper we consider
systems with exogenous inputs.
Moreover, the error bound of this
paper is much tighter than that of \cite{CDC21paper}: 
in contrast to \cite{CDC21paper}, with the growth of the number of observations,
the error bounds of this paper converge either to zero (the one based on 
R\'enyi divergence) or to a constant involving KL-divergence. 
Finally, the proof technique is completely different from that of
\cite{CDC21paper}.
\par
\textbf{Finite-sample bounds for system identification of LTI systems.}
Guarantees for asymptotic convergence of learning algorithms is a classical
topic in system identification \cite{LjungBook}.
Recently, several publications on finite-sample  bounds for learning linear dynamical systems were derived, without claiming completeness \cite{simchowitz2018learning,simchowitz2019learning,simchowitz2021statistical,oymak2021revisiting,lale2020logarithmic,foster2020learning,NEURIPS2018_d6288499,Pappas1,SarkarRD21}. 
First, all the cited papers propose a bound which is valid only for models generated by a specific learning algorithm. In particular, these bounds do not relate the generalization error with the empirical loss for arbitrary models, i.e., they are not PAC(-Bayesian) bounds. This means that in contrast to the results of this paper, the bounds of the cited papers cannot be use for analyzing algorithms others than for which they were derived.
Second, many of the cited papers
do not derive bounds on the infinite horizon prediction error.
More precisely, \cite{oymak2021revisiting,SarkarRD21,lale2020logarithmic,Pappas1,Simchowitz_Foster_2020} provided error bounds for the difference of the first $T$ Markov-parameters of the estimated and true system for a specific
identification algorithm. However, in order to characterize the infinite horizon prediction error,
we need to take $T=\infty$.
For $T=\infty$ the cited bounds become infinite, i.e., vacuous. 
In addition, in contrast to the present paper,  \cite{oymak2021revisiting,SarkarRD21,simchowitz2018learning} deals only with the deterministic part of the stochastic LTI, \cite{Pappas1} deals only with the stochastic part.
Note that the error bounds of
the cited papers
converge to their limit at rate $O(\frac{\ln(N)}{\sqrt{N}})$, which is comparable to the rate $O(\frac{1}{\sqrt{N}})$ of this paper.
This being said, the cited papers 
provide bounds on the parameter estimation error, and many of them allow marginally stable systems.
\par
\textbf{Outline of the paper}
In Section \ref{sect:learning} we define the learning problem. In Section \ref{sect:pac:gen} we describe a PAC-Bayesian framework.
In Section \ref{sec:mainResults} the proposed analytic PAC-Bayesian-Like error bound is showcased and discussed. In Section \ref{sec:numEx} a numerical example is presented.

\section{Problem formulation}
\label{sect:learning}
\textbf{ Notation and terminology} To enhance readability 
We occasionally use $\triangleq$ to denote ''defined by''. 
Let $\F$ denote a $\sigma$-algebra on the set $\Omega$ and $\bP$ be a probability measure on $\F$. Unless otherwise stated all probabilistic considerations will be with respect to the probability space $(\Omega,\F,\bP)$, and we let $\bE(\z)$ denote expectation of the stochastic variable $\z$. 
We typically use bold face letters to indicate stochastic variables/processes. 
Each euclidean space is associated with the topology generated by the 2-norm $\|\cdot\|_2$, and the Borel $\sigma$-algebra generated by the open sets. The induced matrix 2-norm is also denoted $\|\cdot\|_2$. 
We will call a square matrix a Schur matrix, if all its eigenvalues are inside the unit disk. 

A \emph{stochastic linear-time invariant (LTI) systems with inputs in state-space form} \cite[Chapter 17]{LindquistBook} is a dynamical system
of the form
\begin{equation}\label{eq:assumedSys}
	\begin{split}
		\x(t+1)=A\x(t)+B\mathbf{u}(t)+\bnu(t),\\
		\y(t)=C\x(t)+D\mathbf{u}(t)+\etab(t)
	\end{split}
\end{equation}
defined for all $t \in \mathbb{Z}$, where 
$A,B,C,D$ are $n \times n$, $n \times n_u$, $n_y \times n$ and $n_y \times n_u$ matrices respectively , $A$ is a Schur matrix, $\bnu,\etab$ are zero-mean Gaussian i.i.d processes,
$\mathbf{u}$, $\x$, are zero-mean stationary Gaussian processes, $\mathbf{u}(t)$ and $\begin{bmatrix} \etab^T(t),\bnu^T(t) \end{bmatrix}^T$ are independent, and $\x(t)$ and $\begin{bmatrix} \bnu^T(t),\etab^T(t) \end{bmatrix}^T$ are independent. 
The process $\x$ is called the state process, and
$\bnu$ is called the  process noise and $\etab$ is the 
measurement noise. 
If $B,D$ are absent from \eqref{eq:assumedSys}, then 
we say that \eqref{eq:assumedSys} is an 
\emph{autonomous stochastic LTI system}
\par
Loosely speaking, the  learning problem for stochastic LTI systems is as follows:
based on a finite number of samples $\{y(t),u(t)\}_{t=1}^{T}$  of $\{\y(t),\mathbf{u}(t)\}_{t=1}^{T}$, estimate the matrices $A,B,C,D$
of \eqref{eq:assumedSys}. In addition, often the variance of the noises $\bnu$,$\etab$ is also estimated. 
This naive problem formulation is not satisfactory. First, the problem is not well-posed, as the matrices of \eqref{eq:assumedSys} are not uniquely determined by the
input and output process, not even up to a linear state-space transformation
\cite[Chapter 17]{LindquistBook}. Second, this formulation does not explicitly involve the
prediction error. 
For these reasons, for learning 
\cite{LjungBook,simchowitz2019learning,lale2020logarithmic}  stochastic LTI
systems are commonly viewed as devices for predicting current outputs based on past and current inputs and possibly past outputs. Then  the learning problem of stochastic LTI systems is recast as learning a good (optimal) predictor. 
If only inputs are used for prediction, then 
only the matrices $A,B,C,D$ can be estimated.
If past outputs are also involved, then
the covariances of noises $\etab,\bnu$ can also be
estimated.
\par
In the rest of the section we, state the problem of learning predictors realized by LTI systems (Subsection \ref{sect:learn}), and describe
how to interpret stochastic LTI systems as predictor (Subsection \ref{sect:LTIaspred}).
%
\subsection{The problem of learning predictors realizable by LTI systems}
\label{sect:learn}
In order to define the learning problem, we have to specify the space of features and labels, our assumptions on the data generator, and a set of predictors (hypotheses). 
\par
\textbf{Labels and features}
We use the following notation, $\mathcal{Y}\triangleq\mathbb{R}^{n_y}$, $\mathcal{W}\triangleq\mathbb{R}^{n_w}$ 
and for the disjoint union $\mathcal{W}^{*}\triangleq\bigsqcup_{k=1}^{\infty} \mathcal{W}^k$.
We write $w=(w_1,\ldots,w_k)$ 
for an element in $\mathcal{W}^{*}$. \par
Intuitively, the space $\mathcal{W}^{*}$ will be our feature space, and $\mathcal{Y}$ will be our set of labels. 
In general, a predictor (hypothesis) is a map
$f:\mathcal{W}^{*} \rightarrow \mathcal{Y}$ from the feature space to the space of labels.
The set  $\mathcal{Y}$ will be the set of currents outputs of a 
stochastic LTI we would like to learn. The set 
$\mathcal{W}^*$ will be the set of values used for prediction. 
\par
\textbf{Data generator}
Let us fix a stochastic process $\y$ taking values in $\mathcal{Y}$, and a stochastic process $\mathbf{u}$ taking values in $\mathcal{U}\triangleq \mathbb{R}^{n_u}$. 
These stochastic processes share time axis $\mathbb{Z}$, that is, for any $t\in\mathbb{Z}$, $\y(t):\Omega\to\mathcal{Y};\omega\mapsto\y(t)(\omega)$, and $\mathbf{u}(t):\Omega\to\mathcal{W};\omega\mapsto\mathbf{u}(t)(\omega)$
are random vectors on $(\Omega,\bP,\F)$. 
Moreover, 
$\y(t)$ will correspond to outputs  and $\mathbf{u}$ will correspond to inputs of the stochastic LTI systems we would like to learn. Moreover, in order to unify the notation, we will
introduce the process $\w$ taking values in $\mathcal{W}$, where either
\par
$\bullet~$\emph{$\w(t)=\mathbf{u}(t)$, $n_w=n_u$, $\mathcal{W}=\mathcal{U}$} in which case the to be learnt models will predict current outputs  based on past and current inputs, or \par
$\bullet~$\emph{$\w(t)=\begin{bmatrix} \y^T(t) & \mathbf{u}^T(t) \end{bmatrix}^T$, $n_w=n_y+n_u$, $\mathcal{W}=\mathbb{R}^{n_y+n_u}$}, in which case, the to be learnt models will use past outputs in addition to current and past outputs to predict the current output.
\par
In the learning problem of this paper, our training data 
 $\mathcal{S}_N\triangleq\{y(t),u(t)\}_{t=0}^{N}$ will be a sample
 of $\{\y(t),\mathbf{u}(t)\}_{t=0}^{N}$, that is, for some $\omega \in \Omega$, $y(t)=\y(t)(\omega)$ and $w(t)=\w(t)(\omega)$. For the sake of simplicity, sometimes we will use the notation
 $w(t)=\begin{bmatrix} y(t) & u(t) \end{bmatrix}$. 
%
Moreover, we will make the following assumption on the data generator. 
\begin{Assumption}\label{as:generator}
	Let $\y(t)$ and $\mathbf{u}(t)$ be generated by a stochastic LTI system
	\begin{subequations}\label{eq:generator}
		\begin{align}
			\x(t+1) =A_g\x(t)+K_g\e_g(t), \\
			\begin{bmatrix}
				\y(t)\\
				\mathbf{u}(t)
			\end{bmatrix}
			=C_g\x(t)+\e_g(t)
		\end{align}
	\end{subequations}
	where $A_g \in \mathbb{R}^{n \times n},K_g \in \mathbb{R}^{n \times m},C_g \in \mathbb{R}^{m \times n}$ for $n > 0$, $m=n_y+n_u\geq2$ and $\x$, $\y$ and $\e_g$ are stationary, zero-mean, and jointly Gaussian stochastic processes.
	Furthermore, we require that $A_g, A_g-K_gC_g$ to be Schur (all its eigenvalues are inside the open unit circle) and that 
	$\e_g(t)$ is white noise and uncorrelated with $\x(t-k)$, and that $\e_g$ is the innovation process (see \cite{LindquistBook} for definition) of $\begin{bmatrix} \y^T & \mathbf{u}^T \end{bmatrix}^T$.
\end{Assumption}
Note that if there is no feedback from $\y$ to $\mathbf{u}$ (see \cite[Definition 17.1.1]{LindquistBook}),
then, by \cite{LindquistBook,eringis2021optimal}, Assumption \ref{as:generator} is equivalent to 
the existence of a stochastic LTI 
\eqref{eq:assumedSys} with output $\y$ and input $\mathbf{u}$.\par
\textbf{Note:} For learning a predictor, we have the training data set $\mathcal{S}_N$, but we have no knowledge of the matrices $A_g,K_g,C_g$ and noise process $\e_g$. The system \eqref{eq:generator} only defines the assumptions on the data generating process. \par
\textbf{Class of predictors (hypotheses):  predictors realizable by LTI systems}
In this paper we will be interested  in predictors  (hypotheses) 
which arise from linear systems and which are defined below.  
A function $f:\mathcal{W}^*\to\mathcal{Y}$  is said to be realized by a
\emph{linear-time invariant deterministic (LTI) dynamical system}, if there exists and
integer $n$ and matrices 
$\hat{A} \in \mathbb{R}^{n \times n}, \hat{C} \in \mathbb{R}^{n_y \times n}, \hat{B} \in \mathbb{R}^{n \times n_w}, \hat{D} \in \mathbb{R}^{n_y \times n_w}$ such that 
$\hat{A}$ is Schur (all its eigenvalues are inside the unit disk), and for 
all $w_0,\ldots, w_t \in \mathcal{W}$,
	\begin{subequations}\label{eq:predictor}
		\begin{align}
			\hat{\x}(t+1)&=\hat{A}\hat{\x}(t)+\hat{B}w_t, ~ \hat{\x}(0)=0 \\
			f(w_0,\dots,w_t)&=\hat{C}\hat{\x}(t)+\hat{D}w_t
		\end{align}
	\end{subequations}
We will identify the system \eqref{eq:predictor}
with the tuple 
$(\hat{A},\hat{B},\hat{C},\hat{D})$.
We will often denote the predictor 
realizable by the LTI system $(\hat{A},\hat{B},\hat{C},\hat{D})$ by
$f_{(\hat{A},\hat{B},\hat{C},\hat{D})}$.
In this paper, we will be interested in the following hypothesis class.
\begin{Assumption}[Parameterised hypothesis class]\label{as:parameterisation}
	The hypothesis class $\mathcal{F}$ is a parametrized set of LTI predictors   
	$$\mathcal{F}=\{f_{(\hat{A}(\theta),\hat{B}(\theta),\hat{C}(\theta),\hat{D}(\theta))} \mid\theta\in\Theta\}$$
	with $\Theta$ a compact set, and  
	$\hat{A}(\theta)$,$\hat{B}(\theta),\hat{C}(\theta),\hat{D}(\theta)$ continuous functions of 
	$\theta$ taking values in the sets of 
	$\hat{n}\times\hat{n}$, $\hat{n}\times n_w$,
	$n_y\times \hat{n}$ and $n_y\times n_w$
	matrices respectively.
	 Furthermore, we assume that for any $\theta$, $\hat{A}(\theta)$ 
	 is a  Schur matrix, and
	 if $\w(t)=\begin{bmatrix} \y^T(t), & \mathbf{u}^T(t) \end{bmatrix}^T$, then
	 $\hat{D}(\theta)=\begin{bmatrix} 0 & \hat{D}_u(\theta) \end{bmatrix}$ for some $n_y \times n_u$ matrix $\hat{D}_u(\theta)$, i.e., $\hat{D}(\theta)\w(t)$ depends only on 
$\mathbf{u}(t)$
	 \footnote{The latter assumption is necessary, since otherwise we would be using 
	 the components of $\y(t)$ to predict $\y(t)$, which is not meaningful.}.
\end{Assumption}
Under this assumption, we can use probability densities on the set of predictors $\mathcal{F}$.
The latter will be essential for using the PAC-Bayesian framework. 
Next, we define the notions of empirical and generalization loss for 
predictors which are realized by LTI systems.
\begin{Assumption}[Quadratic loss function]\label{as:lossfunc} \;\\
	We will consider \emph{quadratic loss functions}
	$\ell : \mathcal{Y}\times \mathcal{Y} \ni (y,y^{'}) \mapsto \|y-y'\|_2^2=(y-y')^T(y-y') \in [0,\infty)$.
\end{Assumption}
The empirical loss  of a 
predictor for the data $\{\y(t),\w(t)\}_{t=0}^{N}$
is defined as follows:
we define the random variable
\[\hyf(t\mid s) \triangleq f(\w(s),\ldots,\w(t))\] 
which represents the predicted label for the feature vector formed by the random variables
$\w(s),\ldots,\w(t)$.
	The \emph{empirical loss for a predictor $f$ }and processes $(\y,\w)$ is defined by
	\begin{equation}
		\hat{\mathcal{L}}_{N}(f)\triangleq\frac{1}{N}\sum_{i=0}^{N-1} \ell(\hyf(i \mid 0), \y(i)).\label{eq:EmpLoss}
	\end{equation}
The definition of the generalization error is a 
bit more involved. Namely, we are using varying number of inputs for predictions and 
hence 
the expectation $\bE[\ell(\hyf(t \mid 0),\y(t))]$ depends on $t$.
This will hold true even if the processes $\y$ and $\w$ are stationary.
Note that this issue is specific for state-space models:  autoregressive models 
always use the same number of inputs to make a prediction, see Remark \ref{remark:ARMAdifference}.
In this paper we will opt for looking at the case when the size of the past used for the prediction is infinite.
To this end, we need the following result.
\begin{lemma}[Infinite past prediction]\label{l:ihp}
    \; \\
    The limit 
	\( \hyf(t)=\lim_{s \rightarrow -\infty} \hyf(t \mid s) \)
	exists
	in the mean-square sense for all $t$, the process $\hyf(t)$ is stationary, and
	\(	\bE[\ell(\hyf(t),\y(t))]=\lim_{s \rightarrow -\infty} \bE[\ell(\hyf(t \mid s),\y(t))] \)\cite{HannanBook}.
\end{lemma}
This motivates us to introduce the following definition.
	The quantity
	\[ \mathcal{L}(f)=\bE[\ell(\hyf(t),\y(t)]=\lim_{s \rightarrow -\infty} \bE[\ell(\hyf(t \mid s),\y(t))]
	\]
	is called the \emph{generalization loss} of the predictor $f$ when applied to
	process $(\y,\w)$.
\par
Intuitively, $\hyf(t)$ can be interpreted as the prediction of
$\y(t)$ generated by the predictor $f$ based on all (infinite) past and present values of $\w$. As stated in Lemma~\ref{l:ihp} we consider the special case when $\hyf(t)$ is the mean-square limit of $\hyf(t \mid s)$ as
$s \rightarrow -\infty$. Clearly, for large enough $t-s$, the empirical loss, is close to the generalization loss. 
In fact, it is standard practice in learning dynamical systems \cite{LjungBook} to use $\mathcal{L}(f)$ as the measure of fitness
of the predictor.
With these definitions in mind, the learning problem considered in this paper can be stated as follows.
\begin{Problem}[Learning problem]
\label{learn:prob}
Compute a predictor $\hat{f} \in \mathcal{F}$ from a sample 
$\mathcal{S}_N=\{y(t),w(t)\}_{t=0}^{N}$ 
of the random variables $\{\y(t),\w(t)\}_{t=0}^{N}$ 
such that  the generalization loss $\mathcal{L}(\hat{f})$ is small.
\end{Problem}
\begin{Remark} \label{remark:ARMAdifference}
It is known
\cite[Section 4.2]{LjungBook} that the LTI system 
\eqref{eq:predictor} can be
rewritten as an ARX model: 
\begin{equation}
\label{lti:arx}
\hat{\y}_f(t|s)=\sum_{i=1}^{n}\hat{\gamma}_i \hat{\y}_f(t-i|s) + \sum_{i=0}^{n-1}\hat{\eta}_i \w(t-i)
\end{equation}
At a first glance this is similar to classical ARX predictors, where
\(\hat{\y}(t)=\sum_{k=1}^n\hat{\alpha}_k \y(t-k) + \sum_{i=0}^{n-1}\hat{\beta}_i \w(t-i) \)
where $\y$ is predicted based on the last $n$ values $\y$ and $\w$.
However, in contrast to classical ARX models, in \eqref{lti:arx} we do not use the past values of $\y$, but the past values
of the prediction $\hat{\y}_f$.
This difference has significant consequences, in particular, it means that
the previous results \cite{shalaeva2019improved} do not apply.  Note that \cite{alquier2012pred,alquier2013prediction} studied autoregressive models
without inputs (nonlinear AR models), 
so those results are not applicable either.
In fact, the problem of learning LTI systems with inputs, or, which is almost equivalent, learning LTI predictors,  is essentially equivalent to learning ARMA models, and the latter is much more involved than learning
ARX models.
\end{Remark}
\subsection{Relationship between Problem \ref{learn:prob} and system identification of stochastic LTI systems}
\label{sect:LTIaspred}
In order to relate the learning problem from Problem \ref{learn:prob}
with the intuitive 
formulation of the system identification problem for stochastic LTI systems,
we associate predictors with stochastic LTIs. The discussion below is based on
\cite{LjungBook,LindquistBook}.
\par
Assume that \eqref{eq:assumedSys}  is in the innovation form,i.e.,
\begin{equation}
\label{eq:assumedSys:innov}
	\begin{split}
	&	\x(t+1)=A\x(t)+B\mathbf{u}(t)+K\e(t),\\
	&	\y(t)=C\x(t)+D\mathbf{u}(t)+\e(t)
	\end{split}
\end{equation}
where 
$A-KC$ is a Schur matrix, and $\e(t)$ is the so called
innovation process, i.e., 
$\e(t)=\y(t)-\hat{\y}(t)$, where $\hat{\y}(t)$ is the best (minimum variance)  linear prediction
of $\y(t)$ based on the past outputs and on the past and current inputs. 
\par
It is well-known (see \cite[Chapter 17]{LindquistBook}), that if that there is no feedback from $\y$ to $\mathbf{u}$ (see \cite[Chapter 17]{LindquistBook} for the definition of this notion),   and
Assumption \ref{as:generator} holds, then \eqref{eq:assumedSys} can always
be transformed to \eqref{eq:assumedSys:innov}.
We can associate a predictor with \eqref{eq:assumedSys:innov} in two manners. 
The  first one is to ignore the presence of noise in \eqref{eq:assumedSys:innov} and associate with \eqref{eq:assumedSys:innov} the predictor 
$f=f_{(A,B,C,D)}$.  In this case, $\mathcal{W}=\mathbb{R}^{n_u}$ and 
$\w(t)=\mathbf{u}(t)$, $t \in \mathbb{Z}$. 
It then follows that the infinite-past prediction $\hat{\y}_f(t)$ of the predictor $f=f_{(A,B,C,D)}$
relates to $\y$ as follows:
\(
   \y(t) =\hat{\y}_f(t)+ \y^s(t) 
\)
 where 
the process $\y^s(t)$ depends only on the noise and it 
represents the inherent (smallest variance) error  when trying to predict
$\y(t)$ based on $\{\mathbf{u}(s)\}_{s \le t}$. 
If there is no feedback from $\y$ to $\mathbf{u}$, then
\emph{the predictor $f=f_{(A,B,C,D)}$ is the best (minimum variance) linear prediction  of $\y(t)$ based on $\{\mathbf{u}(s)\}_{s \le t}$. }
\par
The approach above does not allow us to estimate the noise gain $K$ and the stochastic noise $\e$. 
To accomplish the latter, predictor needs to depend on past of $\y$, see \cite{Katayama:05}. In this case, $\mathcal{W}=\mathbb{R}^{n_y+n_u}$, 
$\w(t)=\begin{bmatrix} \y^T(t) & \mathbf{u}^T(t) \end{bmatrix}$, and the 
corresponding predictor $f=f_{(\hat{A}_0,\hat{B}_0,\hat{C}_0,\hat{D}_0)}$ associated with
\eqref{eq:assumedSys:innov} will be such that
$\hat{B}_0=\begin{bmatrix} K & B-KD \end{bmatrix}$, $\hat{C}_0=C$,
$\hat{D}_0=\begin{bmatrix} 0 & D \end{bmatrix}$, $\hat{A}_0=A-KC$.
Indeed, in this case, by 
using $\e(t)=\y(t)-C\x(t)-D\mathbf{u}(t)$ we can rewrite \eqref{eq:assumedSys:innov}
as $\x(t+1)=\hat{A}_0\x(t)+\hat{B}_0\w(t)$,
and hence 
$\hat{\y}_f(t)=\hat{C}_0\x(t)+\hat{D}_0\w(t)=C\x(t)+D\mathbf{u}(t)=\y(t)-\e(t)$
is the best linear prediction of $\y(t)$
based on past outputs and past and current inputs.
To sum up, \emph{we can associate \eqref{eq:assumedSys:innov} with the predictor $f_{(\hat{A}_0,\hat{B}_0,\hat{C}_0,\hat{D}_0)}$, which acts on the past and current inputs and past outputs, and which generates the smallest prediction error.} 
\par
That is, 
with each stochastic LTI  in innovation form \eqref{eq:assumedSys:innov} we can associate two types of optimal (smallest variance) linear predictors: the first one uses only past and current inputs to predict the output, the other one uses past outputs too. 
\par
This then means that the problem of estimating an LTI system boils down to
solving Problem \ref{learn:prob}.
More precisely, a solution to Problem \ref{learn:prob}  for $\w=\mathbf{u}$ respectively  $\w=\begin{bmatrix} \y^T & \mathbf{u}^T \end{bmatrix}^T$ allows us to estimate the matrices $A,B,C,D$  respectively $A,B,C,D,K$ of \eqref{eq:assumedSys:innov}, 
by identifying them with the matrices of the corresponding
predictor.  Moreover, the estimates of
the matrices $A,B,C,D,K$ can be used to estimate the 
covariance of the innovation noise $\e(t)$, see   \cite[Chapter 9, page 260]{Katayama:05}.
This identification is theoretically justified under suitable minimality assumptions
on the underlying system and the predictor space. For a more detailed discussion see Appendix C of the supplementary materials. 

\section{PAC-Bayesian Framework}
\label{sect:pac:gen}
Below we present the adaptation of the PAC-Bayesian framework for LTI systems.
To this end, let $B_{\Theta}$ be the $\sigma$-algebra of Lebesque-measurable subsets of the parameter set $\Theta\subset\mathbb{R}^d$, and $m$ denote the Lebesque measure on $\mathbb{R}^d$. With the identification $\theta \leftrightarrow f_\theta=f$ in mind we then define
\begin{equation}
	\underset{f \sim \rho}{E} g(f)\triangleq\int_{\theta \in \Theta} \rho(\theta)g(f_{\theta})dm(\theta)\label{eq:Edef}
\end{equation}
with $\rho$ a probability density function on the measure space $(\Theta,B_{\theta},m)$, and $g:\mathcal{F} \rightarrow \mathbb{R}$ a map such that
$\Theta \ni \theta \mapsto g(f_{\theta})$ is measurable and absolutely integrable. 
The essence of the PAC-Bayesian approach is to prove 
that for any two densities $\pi$ and $\hat{\rho}$ on $\mathcal{F}$, and
any $\delta\in(0,1]$, 
\begin{align} 
\label{T:pac:gen}
	\bP \Big( \Big\{ \omega \in \Omega  \mid  \forall \hat{\rho}:
	\underset{f\sim \hat{\rho}}{E} \mathcal{L} (f) \le \kappa(\omega)&
	\Big \}\Big) > 1-\delta,
\end{align}
with
$$
\kappa(\omega)=\underset{f\sim \hat{\rho}}{E} \hat{\mathcal{L}}_{N}(f)(\omega) + r_N
$$
and $r_N=r_N(\pi,\hat{\rho},\delta)$ an error term.
We may think of $\pi$ as a prior distribution density function and $\hat{\rho}$ as any candidate to a posterior distribution on the space of predictors. 
The inequality \eqref{T:pac:gen} says that the average generalization error for models sampled from the posterior distribution is smaller than the average empirical loss for the posterior distribution plus the error terms $r_N$. \par
The proposed bounds in this paper are PAC-Bayesian-Like, since our theorems are formulated as "given a specific posterior distribution $\hat{\rho}$" and not more correct "for all posterior distributions $\hat{\rho}$ absolutely continuous w.r.t $\pi$". While our formulation is weaker in theoretical aspect, if one is to use numerical methods to estimate $\hat{\rho}$ by minimising $\kappa(\omega)$, then there is no difference between the two notions.
A learning algorithm
can be thought of as fixing a prior $\pi$ and then
choosing a posterior $\hat{\rho}$ for which
$\kappa(\omega)$ is small.
Moreover, $\kappa(\omega)$ can be viewed as a cost function involving the
empirical loss and the regularization term $r_N$.
The learned model
is either sampled from the posterior density $\hat{\rho}$, or it is chosen
as the one with maximal likelihood w.r.t. $\hat{\rho}$. 
Inequality \eqref{T:pac:gen} then gives guarantees on the 
generalization loss of the learned model.
For more details on using PAC-Bayesian bounds see \cite{alquier2021userfriendly}
For \eqref{T:pac:gen} to be useful, 
the term  $r_N$ 
should converge to a small constant, preferably zero,  as $N \rightarrow \infty$, and to
be decreasing in $\delta$.
There are two major classes of \eqref{T:pac:gen}. 
\paragraph{Error bounds using Kullback-Leibler divergence. } 
The first class uses \cite[Theorem 3]{nips-16}, based on
Kullback-Leibler divergence (KL-divergence for short), thus expressing the error term $r_N$ as:
    \begin{align} 
	r_N^{KL}=
		 \dfrac{1}{\lambda}\!\left[ KL(\hat{\rho} \|\pi) +
	\ln\dfrac{1}{\delta}+ \Psi_{\pi}(\lambda,N)  \right] \label{T:pac},
	\end{align}
where $\lambda > 0$ and
	$KL(\hat{\rho} \mid \pi)\triangleq E_{f\sim\hat{\rho}} \ln \frac{\hat{\rho}(f)}{\pi(f)}$ is the KL-divergence between
	$\pi$ and $\hat{\rho}$, and 
	\begin{equation}
	\label{T:pac:2}
	   \Psi_{\pi}(\lambda,N) \triangleq \ln E_{f\sim\pi} \bE[e^{\lambda(\mathcal{L}(f)-\hat{\mathcal{L}}_{N}(f))}]
	\end{equation}
That is, in this case $r_N^{KL}$
involves the KL-divergence and a free parameter $\lambda$.
The density which minimizes $\kappa(\omega)$, with $r_N^{KL}$ from \eqref{T:pac} is known as the Gibbs-posterior
\cite{alquier2021userfriendly} and it can be explicitly computed.
The disadvantage of this approach is that it is difficult to bound
$\Psi_{\pi}(\lambda,N)$. 
\paragraph{Error bounds using R\'enyi divergence.}
Here we use \cite[Theorem 1]{alquier:hal-01385064}, \cite[Theorem 8]{pacRenyi}, based on R\'enyi divergence, thus expressing the error term $r_N$ as: 
\begin{align}
 r_N^R \ge \delta^{-\frac{1}{r}} D_{r}(\hat{\rho}\mid\mid \pi) \left ( E_{f\sim\pi} \bE[(\mathcal{L}(f)-\hat{\mathcal{L}}_N(f))^r]\right )^\frac{1}{r},\label{eq:general-RenyiBound} 
\end{align}
where $D_{r}(\hat{\rho}\mid\mid \pi)\triangleq\left ( E_{f\sim\pi}\left (\frac{\hat{\rho}(f)}{\pi(f)}\right )^\frac{r}{r-1}\right )^\frac{r-1}{r}$ 
denotes the exponential of R\'enyi $\alpha$-divergence, with $\alpha=\frac{r}{r-1}$.
The advantage of this type of error bounds is that it only requires handling of high-order moments of 
$(\mathcal{L}(f)-\hat{\mathcal{L}}_N(f))$, instead of the whole
moment generating function.
The drawback is that R\'enyi divergence may lead to more conservative
error bounds \cite{pacRenyi}, and that it is difficult to compute $\hat{\rho}$ which minimizes the right-hand
side of \eqref{T:pac:gen} with $r_N^R$ from \eqref{eq:general-RenyiBound}, see \cite{alquier:hal-01385064} for an explicit formula. 

\section{Main Results} \label{sec:mainResults}
In this paper we derive PAC-Bayesian-Like bounds  \eqref{T:pac:gen}
for LTI systems, both using KL- and R\'enyi-divergence.
The main idea is to use the change of measure inequalities from \cite[Theorem 3]{nips-16} and \cite[Theorem 8]{pacRenyi}. The major
challenge is to bound the corresponding moment generating function/higher-order moments of $(\mathcal{L}(f)-\hat{\mathcal{L}}_N(f))$.
For both approaches, we have to deal with the same 
technical issues.  Namely, the processes involved are not i.i.d.. Moreover, they are not bounded, and the loss function is not Lipschitz.
In addition, the empirical loss $\hat{\mathcal{L}}_N(f)$ is not
an unbiased estimate of the generalization loss $\mathcal{L}(f)$. This 
is specific to state-space representations,
for auto-regressive models 
considered in \cite{alquier2012pred,alquier2013prediction,alquier:hal-01385064} this problem does not occur. 
All these issues make it impossible to directly apply existing techniques \cite{alquier2012pred,alquier2013prediction,alquier:hal-01385064}.
%
\par In order to simplify the notation,  we state the main results only for scalar output, i.e. $n_y=1$, the general case is discussed in Remark \ref{rem:mimo} and presented in Appendix B of the supplementary material.\par 
As the first step, we replace the empirical loss $\hat{\mathcal{L}}_{N}(f)$ by
\begin{equation}
	\label{inf:emp:pred}
	V_N(f)\triangleq\frac{1}{N}\sum_{i=0}^{N-1}(\y(i)-\hyf(i))^2
\end{equation}
where the finite-horizon prediction $\hyf(t\mid 0)$ is replaced by the 
infinite horizon prediction $\hyf(t)$ defined in Lemma \ref{l:ihp}.
The advantage of $V_N(f)$ over $\hat{\mathcal{L}}_{N}(f)$
is that $V_N(f)$ is an unbiased estimate of the generalization loss
$\mathcal{L}(f)$, i.e.,
$\bE[V_N(f)]=\mathcal{L}(f), $
hence, usual techniques for deriving error bounds are easier to
extend to $V_N(f)$ than to $\hat{\mathcal{L}}_{N}(f)$.
Moreover, from Lemma B.7 in Appendix B of the supplementary material, it follows that $\hat{\mathcal{L}}_{N}(f)-V_N(f)$ converges to zero as $N \rightarrow \infty$
%
in the mean sense. 
In order to derive upper bounds on the errors of the type \eqref{T:pac} and \eqref{eq:general-RenyiBound}, 
we apply change of measures on $\lambda(\mathcal{L} (f) -  V_N(f))$ instead of $\lambda(\mathcal{L} (f) -  \hat{\mathcal{L}}_{N}(f)))$. Then, to obtain bounds with empirical loss $\hat{\mathcal{L}}_N$, we prove
a probabilistic error bound for $V_N(f)-\hat{\mathcal{L}}_{N}(f)$
which converges to $0$ as $N$ goes to $\infty$. By combining these
two bounds we will finally derive a PAC-Bayesian-Like error bound for
$\hat{\mathcal{L}}_{N}(f)$.
 For every predictor $f$ 
 we define the following constants. 
	\begin{Definition}[Constants $G(f),G_e(f)$]
	\label{def:constants}
	 Let $f=(\hat{A},\hat{B},\hat{C},\hat{D})$ be a predictor.
     Let
	$A_g,K_g,C_g$ be the matrices of the data generator  from Assumption \ref{as:generator}. Define
	the matrices $(A_e,K_e,C_e,D_e)$ as
	  \( D_e=
	          I-\hat{D}_w 
\)
	\begin{align*}
	   & A_e=
	   \begin{bmatrix} A_g & 0 \\ \hat{B}C_w & \hat{A}  \end{bmatrix} 
	   ~~
	   K_e=
	    \begin{bmatrix*}[l] K_g \\
	    \hat{B}_w
	    \end{bmatrix*} 
	     ~ C_e=
	         \begin{bmatrix} (C_1-\hat{D}C_w)^T \\ -\hat{C}^T \end{bmatrix}^T 
	    \end{align*}
	where $C_g=\begin{bmatrix} C_1^T & C_2^T \end{bmatrix}^T$ and $C_1$ has $n_y$ rows and $C_2$ has $n_u$ rows; and
	$(C_w,\hat{B}_w,\hat{D}_w)=(C_2,\begin{bmatrix} 0 & \hat{B} \end{bmatrix}, \begin{bmatrix} 0 & \hat{D} \end{bmatrix})$
	if $\w=\mathbf{u}$, and 
	$(C_w,\hat{B}_w,\hat{D}_w)=(C_g,\hat{B},\hat{D})$, if $\w=\begin{bmatrix} \y^T & \mathbf{u}^T \end{bmatrix}^T$.
	With these definitions, 
  \begin{align*}
	   & G_e(f)\hspace{-2pt}=\hspace{-2pt}\|(A_e,K_e,C_e,D_e)\|_{\ell_1}\hspace{-2pt}\triangleq\hspace{-2pt}\|D_e\|_2 \hspace{-2pt}+\hspace{-2pt}\sum_{k=0}^{ \infty} \|C_eA_e^{k}K_e\|_2 \nonumber \\
	   &  G(f)=G_{-1}(f)G_0(f)G_1(f)G_2(f)G_3(\w ) \\ 
        & G_{-1}(f)= 
         \left(2\sum_{k=0}^{\infty}  \|A_e^k \|_2^2 +4 \right)^{\frac{1}{2}},~
		G_0(f)=   \sum_{k=0}^{\infty} \| 
		\hat{A}^k \|_2 \\
		& G_1(f)=\sqrt{\|D_e\|_2^2 +\sum_{k=0}^{\infty} \|C_e\|_2^2\|A_e^k\|_2^{2}\|K_e\|_2^2}, \\ 
	& G_2(f)=\|\hat{D}\|_2 +\sum_{k=0}^{\infty} \|\hat{C}\|_2\|\hat{A}^k\|_2\|\hat{B}\|_ 2 \\
	& G_3(\w)= \sqrt{\mu_{\max}(Q_e)K_w} 
		\end{align*}
	where $K_w$ is un upper bound of $\|\bE[\w(t)\w^T(t-r)]\|_2$.
	\end{Definition}
The interpretation of the various terms appearing in  Definition \ref{def:constants} is as follows.
\begin{Remark}[Interpretation of constants]
\textbf{Matrices $A_e,K_e,C_e,D_e$}
  These matrices represent the matrices of the LTI system driven
  by the innovation process $\e_g$ of $(\y^T,\w^T)^T$, output
  of which is $\y-\hat{\y}_f$, i.e.,
\begin{equation}
\label{error-sys1}
\begin{split}
   \tilde{\x}(t+1)=A_e\tilde{\x}(t)+K_e\e_g(t), \\
   \y(t)-\hat{\y}_f(t)=C_e\tilde{\x}(t)+D_e\e_g(t)
\end{split}  
\end{equation}
\textbf{The term $G_{-1}(f)$} depends on the predictor and on the
data generating system, and it characterizes the stability of the
error system: if $\rho \in (0,1)$ is the
maximum of moduli of eigenvalues of $\hat{A}$ and $A_g$, then 
$\|A_e^k\|_2 \le C\rho^k$ and $G_{-1}(f) \le \sqrt{2(\frac{C^2}{1-\rho^2}+2)}$. 
\textbf{The term $G_0(f)$} depends only on the predictor $f$, and 
it characterizes the stability of $f$: if the spectral radius of $\hat{A}$
is smaller than $\rho < 1$, then $\|\hat{A}^k\|_2 < K\rho^k$ and 
$G_0(f) \le \frac{K}{1-\rho}$. The constant $\rho$ is the exponent of
the decay of the influences of the initial state on the predictor $f$.
\textbf{The term $G_1(f)$} is an upper bound on the $H_2$ norm  \cite{Katayama:05} 
of the  error system $(A_e,K_e,C_{e},D_e)$ from \eqref{error-sys1} for $n_y=1$. 
Hence, if it is small, then the $H_2$ error of the error system is small too.
Evaluating $G_1(f)$ requires the knowledge of the data generating system and the predictor.
\textbf{The expression $G_2(f)$} is an
upper bound on the $\ell_{\infty}$ norm  \cite{DahlehTACFeedback}
of the predictor system
$(\hat{A},\hat{B},\hat{C},\hat{D})$, and depends only on the predictor $f$. 
\textbf{The term $G_3(\w)$} depends only on the process $\w$; 
$K_w$ is an upper bound on the power spectrum
$\Phi_{\w}(z)=\sum_{k=-\infty}^{\infty} \Lambda_k^{\w} z^{-k}$,
of $\w$: if  $K_w^2I \ge \Phi_w(z)$ for all
$z$ on the unit disk, then $K_w \ge \|\Lambda^{\w}_k\|_2$ for all
$k$. That is $K_w$ is an indicator of the richness of $\w$. The term $\mu_{\max}(Q_e)$ denotes the maximum eigenvalue of the covariance matrix $Q_e$ of forward innovation of the generating system \eqref{eq:generator}, see Lemma 6 in Appendix B. \par
That is, the term $G(f)$ thus depends only on the model $f$ and on the data generating system \eqref{as:generator}, but not on the number of data points $N$. 
\end{Remark}

\begin{Theorem}[Bound using KL-divergence]
	\label{thm:pac:altKL}
	Assume $n_y=1$.
	For any two densities $\pi$ and $\rho$ on $\mathcal{F}$, any $\delta \in (0,1]$, and any 
	$0< \lambda < (3(m+1) \mu_{\max}(Q_e)G_e(\Theta)^2)^{-1}$, with $m=n_y+n_u$, 
	 the following inequality \eqref{T:pacAltKL} holds with probability at least $1-2\delta$
 	\begin{align}
     & E_{f\sim \hat{\rho}} \mathcal{L} (f) \le E_{f\sim \hat{\rho}} \hat{\mathcal{L}}_{N}(f)+\hat{r}_N^{KL} \label{T:pacAltKL}
 \\
	&	 \hat{r}_N^{KL}\hspace{-2pt}=\hspace{-2pt}\frac{2}{\delta N} \underset{f\sim \hat{\rho}}{E} G(f)  \hspace{-2pt}+\hspace{-2pt}\dfrac{1}{\lambda}\!\Big[ KL(\hat{\rho} \|\pi) +\ln\dfrac{1}{\delta}	+ \widehat{\Psi}_{\pi}(\lambda,N) \Big ] \nonumber 
	\\
	    & \widehat{\Psi}_{\pi}(\lambda,N)\hspace{-2pt}=\hspace{-2pt} \ln \left(1+\frac{4}{N}\underset{f\sim\pi}{E}\left[\frac{(m+1)!  
	     \left (3K_\mu(f) \right )^2}{(1-3(m+1)K_\mu(f))}\right] \right)  \nonumber 
	     \\ 
	    & \qquad \qquad  \le \ln \left(1+\frac{4}{N} \frac{(m+1)!\left (3K_\mu(\Theta) \right )^2}{(1-3(m+1)K_\mu(\Theta))} \right) \nonumber
	\end{align}
	with $r_N^{KL}\leq \hat{r}_N^{KL}$, and where $KL(\hat{\rho}||\pi)$ is the Kullback–Leibler divergence, $K_\mu(\Theta)=\lambda\mu_{\max}(Q_e)G_e(\Theta)^{2}$,  $K_\mu(f)=\lambda\mu_{\max}(Q_e)G_e(f)^{2}$, and $G_e(\Theta)\triangleq\sup_{f\in\mathcal{F}}G_e(f)$, and the terms
	$G(f)$ and $G_e(f)$ are as in Definition \ref{def:constants}. 
\end{Theorem}
The proof of Theorem \ref{thm:pac:altKL} and its extension to $n_y > 1$ is presented in Appendix B of the supplementary material. 
The term $\hat{r}_N^{KL}$ \eqref{T:pacAltKL} includes the term $\frac{2}{\delta N}E_{f\sim\rho} G(f)$, which comes from using $V_N(f)$ instead of the empirical loss, and does not relate to any of the terms in $r_N^{KL}$ \eqref{T:pac}.  
%
The term $\hat{r}_N^{KL}$ includes an upper bound 
$\widehat{\Psi}_{\pi}(\lambda,N)$ on $\Psi_{\pi}(\lambda,N)$.
This upper bound is increasing in the maximal eigenvalue of the covariance
of the innovation process $\e_g$ of the data $(\y^T,\w^T)^T$.
Intuitively, the latter is a measure of the inherent variability of
$(\y^T,\w^T)^T$ and it tells us how well the current output
$\y$ can at all be predicted using past inputs. 
In addition, $\widehat{\Psi}_{\pi}(\lambda,N)$
involves the average w.r.t. $\pi$ of an increasing function of $G_e(f)$. The latter is proportional to the generalization loss of $f$. 
That is, the better $\y$ can be predicted based on $\w$
the smaller is $\widehat{\Psi}_{\pi}(\lambda,N)$. 
The term
$G_e(\Theta)=\sup_{f\in\mathcal{F}}G_e(f)$ 
expresses the complexity of the model class, it can be viewed as a 
counterpart of VC-dimension.
%
Finally, $\widehat{\Psi}_{\pi}(\lambda,N)$ converges to zero as $N \rightarrow \infty$, which means that 
$\hat{r}_N^{KL}$ converges to the data-independent constant
$\frac{1}{\lambda}(KL(\hat{\rho} \|\pi) + \ln\dfrac{1}{\delta})$ as $N \rightarrow \infty$.
In order to control the term,
in the literature $\lambda$
is chosen so that $\lambda \rightarrow \infty$ as $N \rightarrow \infty$.
Unfortunately, Theorem \ref{thm:pac:altKL} holds only for small enough $\lambda$, hence this approach cannot work to make 
$\hat{r}_N^{KL}$ converge to zero as $N \rightarrow \infty$.
\\
Nevertheless, the result is still useful, as it
provides non-trivial bounds on the generalization loss.
One could also derive a counterpart of the Gibbs-posterior for \eqref{T:pacAltKL} which
minimizes the right-hand side of \eqref{T:pacAltKL}, see Appendix B of the supplementary material. However, the thus derived
posterior density $\rho(f)$ differs from the classical one by the presence of the term $\frac{2}{N\delta}G(f)$ which converges to zero as $N \rightarrow \infty$.\par 
The drawback 
of Theorem \ref{thm:pac:altKL} is that the error term
$\hat{r}_N^{KL}$ does not converge to
$0$ as $N \rightarrow \infty$, and hence the resulting bound is not tight. 
In order to circumvent this problem, inspired by
\cite{pacRenyi,alquier:hal-01385064}, we will use 
\cite[Theorem 8]{nips-16} to derive
error bounds of the form \eqref{eq:general-RenyiBound}.
\begin{Theorem}[Bound using R\'enyi divergence]
	\label{thm:pac:alt1}
	Assume $n_y=1$. 
	For any two densities $\pi$ and $\rho$ on hypothesis class $\mathcal{F}$, any $\delta \in (0,1]$, any even integer $r\geq 2$, and for integer $m=n_y+n_u$, 
	 the following inequality \eqref{eq:RenyiBound} holds with probability at least $1-2\delta$
    \begin{align} 
      &  E_{f\sim\rho}\mathcal{L}(f) \leq E_{f\sim\rho} \hat{\mathcal{L}}_N(f) + \hat{r}_N^R\label{eq:RenyiBound} 
     \\  & \hat{r}_N^R= \frac{2}{\delta N}E_{f\sim\rho} G(f) + \left (\frac{4}{\delta N}\right )^\frac{1}{r} \Phi(\pi,r) \nonumber 
      \end{align}\begin{align}
        &\Phi(\pi,r)= 3\mu_{\max}(Q_e)\left [(m+r-1)!(r-1)\right]^\frac{1}{r}\\
        &\qquad \qquad \qquad  \cdot D_{r}(\hat{\rho}\mid\mid \pi)\left ( E_{f\sim\pi} G_e^{2r}(f) \right )^\frac{1}{r} \nonumber
    \end{align}
    with $r_N^R\leq \hat{r}_N^R$, and where $D_{r}(\hat{\rho}\mid\mid \pi)\triangleq\left ( E_{f\sim\pi}\left (\frac{\hat{\rho}(f)}{\pi(f)}\right )^\frac{r}{r-1}\right )^\frac{r-1}{r}$ 
    is the exponential of the R\'enyi $\alpha$-divergence, with $\alpha=\frac{r}{r-1}$, and
    $G(f),G_e(f)$ is as in Definition \ref{def:constants}.
\end{Theorem}
The proof of Theorem \ref{thm:pac:alt1} and its extension to $n_y > 1$ is presented in Appendix B of the supplementary material.
The bound $\hat{r}_N^R$ \eqref{eq:RenyiBound} includes  
the term $\frac{2}{\delta N}E_{f\sim\rho} G(f)$,
which represents the difference
between the empirical loss and the quantity $V_N(f)$, and does not relate to any terms in \eqref{eq:general-RenyiBound}.
This extra term converges to zero 
at rate  $O(\frac{1}{N})$.
In contrast to Theorem \ref{thm:pac:altKL}, the error bound
$\hat{r}_N^R$ 
converges to zero as $N \rightarrow \infty$. That is, for large enough $N$, it will give a non-trivial  guarantee on the generalization loss.
The rate of convergence of $\hat{r}_N^R$ is
$O\left (\frac{1}{N^{\frac{1}{r}}}\right )$.
The fastest rate $O\left (\frac{1}{\sqrt{N}} \right )$ is achieved for $r=2$.
This rate is comparable with the results of \cite{alquier:hal-01385064} for autoregressive models However, 
it holds for the state-space case.
It is also comparable for the rate $O(\frac{\ln(N)}{\sqrt{N}})$ of the finite-sample error bounds of \cite{lale2020logarithmic,simchowitz2019learning,SarkarRD21,Pappas1,NEURIPS2018_d6288499}, which apply in a more restricted setting, see the discussion in the introduction. \par
As in Theorem \ref{thm:pac:altKL}, the upper bound $\hat{r}_N^R$ of
Theorem \ref{thm:pac:alt1} is proportional to the maximal eigenvalue $\mu_{max}(Q_e)$ of
the covariance matrix $Q_e$ of the innovation process of $(\y^T,\w^T)^T$. As it was noted
before, this eigenvalue measures the variability of $(\y^T,\w^T)^T$. The larger
$\mu_{max}(Q_e)$ is, the more difficult it is to predict $\y$ based on $\w$. \par 
The term $\Phi(\pi,r)$ is proportional to $E_{f\sim\pi} G_e^{2r}(f)$ which measures the average prediction error w.r.t.
$\pi$ of various models. That is, the more concentrated the prior $\pi$ is around the
the optimal predictor, and the smaller the inherent variablity of $\y$ is, the smaller  $\hat{r}_N^R$ is. \par 
The term $E_{f\sim\pi} G_e^{2r}(f) \le \sup_{f \in \Theta} G_e^{2r}(f)$
can be viewed as a measure of the complexity of the hypothesis class, i.e., as a counterpart of VC-dimension. Such terms are standard in PAC-Bayesian inequalities \cite{alquier:hal-01385064,nips-16, alquier2021userfriendly}.
\begin{Remark}[Extension to $n_y > 1$]
\label{rem:mimo}
If $n_y>1$, then we can decompose the problem into predicting each component of $\y$ separately, and find a PAC-bound for each prediction. Then the probability that all individual PAC-Bounds, i.e. $r_N=\sum_{p=1}^{n_y}r_{N,p}$, holds is $1-2n_y\delta$. 
The extension of the results below to $n_y > 1 $ are presented in Appendix B of the supplementary material.
\end{Remark}

\section{Numerical Example} \label{sec:numEx}
\begin{figure}[ht]
    \centering
    \includegraphics[width=0.99\linewidth]{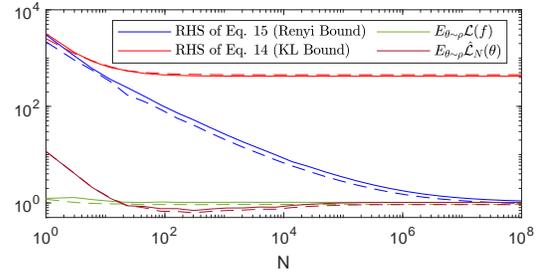}
    \caption{Results of a numerical example, with the
     R\'enyi bound (with $r=2$)  and KL bound ($\lambda= (3(m+1) \mu_{\max}(Q_e)G_e(\Theta)^2)^{-1}=0.005$) , and with $\delta=0.1$ for both cases. For both bounds, the posterior $\rho$ is the Gibbs-posterior related to the KL bound. For details see Appendix A. Solid lines represent the case with $\w=\mathbf{u}$, dashed lines the case with $\w=[\y^T\;\mathbf{u}^T]^T$. }
    \label{fig:fullBound}
\end{figure}
A numerical example has been done, in order to see how conservative the results are, see supplementary material for details and code. We assume that the data is generated by a second order system \eqref{as:generator}, i.e. $n=2$. The hypothesis classes consist of perturbing only one element of $\hat{A}$ matrix, in an interval. 
The prior distribution $\pi$ is the uniform distribution. For the upper bound based on Theorem \ref{thm:pac:alt1}, the posterior is chosen as the gibbs posterior
$ \hat{\rho}(\theta) = Z^{-1} \pi(\theta) e^{-\lambda\hat{\mathcal{L}}_N(\theta)}, $
with $\lambda=10$, and $Z=E_{\theta\sim\pi}e^{-\lambda\hat{\mathcal{L}}_N(\theta)}$. Similarly for the bound based on Theorem \ref{thm:pac:altKL}, gibbs posterior is chosen, however with constant $\lambda= (3(m+1) \mu_{\max}(Q_e)G_e(\Theta)^2)^{-1}=0.005$. In figure \ref{fig:fullBound}, we can see the bound converging with $N$, however due to the restriction on $\lambda$ by Theorem \ref{thm:pac:altKL}, $\lambda$ cannot increase with $N$. Therefore for large $N$, we are left with terms $\lambda^{-1}KL(\rho || \pi)$ and $\lambda^{-1}\ln(\delta^{-1})$, which for small $\lambda$, imply that the bound is relatively big. On the other hand, the bound based on Renyi divergence does converge to empirical loss.

\section{Conclusion}
\label{sect:concl}
In this paper we have 
derived two PAC-Bayesian-Like error bounds
for stochastic LTI systems with inputs. 
The second error bound converges to $0$ as the number of samples $N$ grows with a rate of convergence $O(\frac{1}{\sqrt{N}})$.
Future research will be directed towards extending these results to more general state-space representations and using the results of the paper for deriving oracle inequalities \cite{alquier2021userfriendly}.


\bibliography{bib}
\bibliographystyle{apalike}



\appendix
\onecolumn
\section{Additional details: choice of the posterior and numerical example}
\subsection{Choice of the posterior density and learning }
Below we will discuss how to choose the posterior density when the using
PAC-Bayesian-Like inequalities derived in this paper. \\
If we try to follow the usual procedure for using PAC-Bayesian error bound
in deriving learning algorithms, we will minimize
the right-hand side of the upper bound, i.e., the expression,  $E_{f\sim\rho}\hat{\mathcal{L}}_N(f)+r_N(\pi,\rho,\delta)$ with respect to the posterior density $\rho$? By minimizing the upper-bound, we minimise the generalisation loss.
The posterior density which minimizes the upper-bound then gives rise to a  distribution over predictors. The latter distribution is optimal in the sense that it gives the smallest possible
average generalization loss, where the average is taken over all predictors. 
We can then either randomly sample a predictor from that distribution  or take the 
predictor with the highest likelihood \cite{alquier2021userfriendly}. \\
That is, the goal is to solve the following optimization problem over all the densities on the
set of predictors:
\begin{align}
    \hat{\rho}(f)=\argmin_{\rho} \left(E_{f\sim \rho} \hat{\mathcal{L}}_{N}(f)(\omega) + r_N(\pi,\rho,\delta) \right).
\end{align}
For $r_N(\pi,\rho,\delta)$ of the form \eqref{T:pac}, the optimisation problem becomes
\begin{align}
    \hat{\rho}(f)=\argmin_{\rho} \left (E_{f\sim \rho} \hat{\mathcal{L}}_{N}(f)(\omega) + \frac{1}{\lambda}KL(\rho||\pi) \right). 
\end{align}
In this case, the classical  Donsker \& Varadhan variational formula \cite{alquier2021userfriendly} can be applied, to obtain the following analytical solution,
\begin{align}
    \hat{\rho}(f)=Z^{-1}(\omega)\pi(f) e^{-\lambda\hat{\mathcal{L}}_N(f)(\omega)}, \label{eq:OLDgibbs}
\end{align}
where $Z(\omega)=E_{f \sim \pi} e^{-\lambda\hat{\mathcal{L}}_N(f)}$ is the normalization term.
In the literature, the density \eqref{eq:OLDgibbs} referred to as the Gibbs posterior
\cite{alquier2021userfriendly}. \\
For  the PAC-Bayesian bound proposed in theorem \ref{thm:pac:altKL}, the corresponding optimisation problem differs slightly, by the presence of the term $\frac{2}{\delta N}G(f)$, i.e.
\begin{align}
    \hat{\rho}_G(f)=\argmin_{\rho} \left( E_{f\sim \rho}\left ( \hat{\mathcal{L}}_{N}(f)(\omega)+\frac{2}{\delta N}G(f)\right ) + \frac{1}{\lambda}KL(\rho||\pi) \right).
\end{align}
In this case, the Donsker \& Varadhan variational formula \cite{alquier2021userfriendly} still applies, and it yields the optimal posterior as
\begin{align}
    \hat{\rho}_G(f)=Z_{alt}^{-1}(\omega)\pi(f) e^{-\lambda(\hat{\mathcal{L}}_N(f)(\omega)+2(\delta N)^{-1}G(f))} \label{eq:newgibbs}
\end{align}
where $Z_{alt}(\omega)=E_{f \sim \pi} e^{-\lambda(\hat{\mathcal{L}}_N(f)+2(\delta N)^{-1}G(f))}$ is the normalization term.
Unfortunately, evaluation of the posterior \eqref{eq:newgibbs} requires knowledge of the generating system, due to the presence of $G(f)$. However, the influence of the term $2(\delta N)^{-1}G(f)$ on the posterior \eqref{eq:newgibbs} decays at the rate of $O(\frac{1}{N})$. \\
For the Renyi based bound from theorem \ref{thm:pac:alt1}, minimizing the upper bound w.r.t. the posterior density results in the
following optimization problem:
 \begin{align}
    \hat{\rho}_R(f)=\argmin_{\rho} \left( E_{f\sim \rho}\left ( \hat{\mathcal{L}}_{N}(f)(\omega)+\frac{2}{\delta N}G(f)\right ) + \frac{2}{\sqrt{\delta N}}\Phi(\pi) \right), \label{eq:minRenyi}
\end{align}
with $\Phi(\pi)=3\mu_{\max}(Q_e) \sqrt{(m+1)!}\sqrt{ E_{f\sim\pi}\left (\frac{\hat{\rho}(f)}{\pi(f)}\right )^2}\sqrt{ E_{f\sim\pi} G_e^{4}(f) }$.
 However, there is no easily computable expression for $\hat{\rho}(f)$ from \eqref{eq:minRenyi}, to the best of the authors' knowledge. Therefore, in the next section we will use the more classical Gibbs posterior \eqref{eq:OLDgibbs}, as the posterior density, even though \eqref{eq:OLDgibbs} does not minimise either the KL based bound of theorem \ref{thm:pac:altKL}, or the R\'enyi based bound from theorem \ref{thm:pac:alt1}.  
 
\subsection{Numerical Example}
In this section we will explore a simple toy example, to illustrate theorems \ref{thm:pac:altKL} and \ref{thm:pac:alt1}. The code which generates the figures in this section, can be found in the supplementary material. Throughout this section, we will know what the generating system is, and therefore we will be able to compute the PAC-Bayesian-Like upper bounds proposed in the main text. Firstly, assume that the data is generated by the LTI system in forward innovation form as
\begin{subequations} \label{numEx:fg}
    \begin{align}
        \x(t+1)&=\underbrace{\begin{bmatrix} 0.16& -0.3\\0& -0.05\end{bmatrix}}_{A_g} \x(t) + \underbrace{\begin{bmatrix}0.33 & -0.75 \\ 0 & -0.09\end{bmatrix}}_{K_g} \e_g(t)\\
        \begin{bmatrix} \y(t) \\ \mathbf{u}(t) \end{bmatrix} &= \underbrace{\begin{bmatrix} 1 & 1 \\ 0 & 1\end{bmatrix}}_{C_g} \x(t) + \begin{bmatrix} 1 & 0 \\ 0 & 1\end{bmatrix} \e_g(t)\\
        Q_e=\bE[\e_g(t)\e_g^T(t)]&=\begin{bmatrix} 0.9 & 0.3 \\ 0.3 & 4.15\end{bmatrix}
    \end{align}
\end{subequations}

That is, \eqref{numEx:fg} corresponds to the data generating system from assumption \ref{as:generator} of
the paper.
Note that in this case the dimension $m$ of the innovation process $\e_g$ is $2$, 
as $\y\in \reals^1$ and $\mathbf{u} \in \reals^1$. With data generator in place, now we need to define the hypothesis class, for that we will find initial predictors and then parameterise them.
Since, for the example we have chosen $\mathbf{u}$ to be feedback free of $\y$, we can use the results of \cite{eringis2021optimal}, and from \eqref{numEx:fg} we can obtain the realisation of $\y$ as
\begin{subequations} \label{eq:y_realisation}
	\begin{align}
		\bar{x}(t+1)&=\underset{\tilde{A}}{\underbrace{\begin{bmatrix}A_{1,1}&A_{1,2}-K_{1,2}C_{2,2}-K_{1,1}D_0C_{2,2}\\0&A_{2,2}-K_{2,2}C_{2,2}  \end{bmatrix}}}\bar{x}(t)+\underset{\tilde{K}_u}{\underbrace{\begin{bmatrix}K_{1,2}+K_{1,1}D_0\\K_{2,2}\end{bmatrix}}}\mathbf{u}(t)+\underset{\tilde{K}_y}{\underbrace{\begin{bmatrix} K_{1,1} \\ 0 \end{bmatrix}}}\mathbf{e}_s(t), \\
		\y(t) &= \underset{\tilde{C}}{\underbrace{\begin{bmatrix} C_{1,1} & C_{1,2}-D_0C_{2,2} \end{bmatrix}}}\bar{x}(t) +D_0\mathbf{u}(t) + \mathbf{e}_s(t), 
	\end{align}
\end{subequations}
with
\begin{align}
    \mathbf{e}_s(t)&=\y-\bE[\y\mid \{\y(s)\}_{s<t},\{\mathbf{u}(s)\}_{s\leq t}] \label{eq:es}\\
    D_0&=Q_{e_{1,2}}/Q_{e_{2,2}}\\ 
    A_g&=\begin{bmatrix}A_{1,1}&A_{1,2}\\0&A_{2,2}\end{bmatrix},\quad K_g=\begin{bmatrix}K_{1,1} & K_{1,2}\\0 & K_{2,2}\end{bmatrix},\quad C_g=\begin{bmatrix} C_{1,1} & C_{1,2}\\0 & C_{2,2} \end{bmatrix}.
\end{align}

Now from \eqref{eq:y_realisation} we can obtain the initial predictors for two cases: the first when we wish to predict $\y$ from $\mathbf{u}$, and the second when we wish to predict $\y$ from $\mathbf{u}$ and past of $\y$.

For the first case, we use the feature process $\w=\mathbf{u}$, and the optimal predictor\cite{eringis2021optimal} of $\y(t)$ based on past and present $\mathbf{u}$ is given by 
\begin{subequations}
\begin{align}
    \hat{x}(t+1)=\tilde{A}\hat{x}(t)+\tilde{K}_u\mathbf{u}(t)=&\begin{bmatrix} 0.16& 0.42\\0& 0.04\end{bmatrix} \hat{x}(t)+\begin{bmatrix} -0.72\\-0.09\end{bmatrix} \w(t)\\
    \hat{y}(t)=\tilde{C}\hat{x}(t)+D_0\mathbf{u}(t) =& \begin{bmatrix} 1 & 0.92 \end{bmatrix} \hat{x}(t) + 0.07\w(t)
\end{align}
\end{subequations}
For the second case, we use the feature process $\w=[\y^T, \mathbf{u}^T]^T$, and we can use \eqref{eq:y_realisation} with \eqref{eq:es} and $\hat{y}=\bE[\y\mid \{\y(s)\}_{s<t},\{\mathbf{u}(s)\}_{s\leq t}]$ to obtain 
\begin{subequations}
\begin{align}
    \hat{x}(t+1)=(\tilde{A}-\tilde{K}_y\tilde{C})\hat{x}(t)+\begin{bmatrix}\tilde{K}_y & \tilde{K}_u\end{bmatrix}\begin{bmatrix}\y(t) \\\mathbf{u}(t) \end{bmatrix} =&\begin{bmatrix} -0.17& 0.12\\0& 0.04\end{bmatrix} \hat{x}(t)+\begin{bmatrix} 0.33 & -0.72\\0 & -0.09\end{bmatrix} \w(t)\\
    \hat{y}(t)=\tilde{C}\hat{x}(t)+\begin{bmatrix} 0 & D_0\end{bmatrix}\begin{bmatrix} \y(t) \\ \mathbf{u}(t) \end{bmatrix}=& \begin{bmatrix} 1 & 0.92 \end{bmatrix} \hat{x}(t) + \begin{bmatrix}0& 0.07\end{bmatrix}\w(t)
\end{align}
\end{subequations}

For the purposes of this example and figures we will assume that most of the optimal predictor is known except for one entry. Hence the hypothesis classes for the two different cases are defined as
    \begin{align*}\label{numEx:params}
        \mathcal{F}_1=\Big \{\hspace{-3pt} \left . \Sigma(\theta)\hspace{-2pt}=\hspace{-2pt}\left ( \begin{bmatrix} \theta& 0.43\\0& 0.04\end{bmatrix}\hspace{-2pt},\begin{bmatrix} -0.72\\-0.09 \end{bmatrix}\hspace{-2pt},\begin{bmatrix} 1& 0.92 \end{bmatrix}\hspace{-2pt},\begin{bmatrix} 0.07 \end{bmatrix} \right ) \right | -0.5\leq \theta \leq 0.5 \Big \}\\
        \mathcal{F}_2=\Big \{\hspace{-1pt}  \Sigma(\theta)\hspace{-2pt}=\hspace{-2pt}\Big ( \begin{bmatrix} \theta& 0.12\\0& 0.04\end{bmatrix}\hspace{-2pt},\begin{bmatrix} 0.33 & -0.73\\0&-0.09 \end{bmatrix}\hspace{-2pt},\begin{bmatrix} 1& 0.92 \end{bmatrix}\hspace{-2pt},\begin{bmatrix}0& 0.07 \end{bmatrix} \Big ) \Big | -0.5\leq \theta \leq 0.5 \Big \}
    \end{align*}
Together with the hypothesis class we also need to define the prior distribution over parameters.
For the purposes of the example we will use the uniform distribution as prior. 
\begin{align}
    \pi(\theta)= \mathcal{U}(0,0.4) = \begin{cases}1&,-0.5\leq \theta \leq 0.5\\0&,\text{else}\end{cases}
\end{align}

For the posterior distribution $\hat{\rho}$ we will use Gibbs posterior \eqref{eq:OLDgibbs}, with $\lambda=10$ for theorem \ref{thm:pac:alt1} (R\'enyi bound) and we will use $\lambda= (3(m+1) \mu_{\max}(Q_e)G_e(\Theta)^2)^{-1}=0.005$ for theorem \ref{thm:pac:altKL} (KL bound),
\begin{align*}
	\hat{\rho}(\theta)&=Z^{-1}\pi(\theta)e^{-\lambda\hat{\mathcal{L}}_N(\theta)}\label{numEx:gibbs}\\
	Z&=E_{\theta\sim\pi}e^{-\lambda\hat{\mathcal{L}}_N(\theta)}
\end{align*}
In order to illustrate theorems \ref{thm:pac:altKL} and \ref{thm:pac:alt1}, we need to compute the following quantities: 
\begin{itemize}
    \item $E_{\theta\sim \hat{\rho}} \hat{\mathcal{L}}_{N}(\theta)$
    \item $\frac{2}{\delta N} E_{\theta\sim \hat{\rho}} G(\theta)$
    \item $\hat{\Psi}_{\pi}(\lambda,N)$
    \item $\Phi(\pi,r)$
    \item $E_{\theta\sim \hat{\rho}} \mathcal{L}(\theta)$
\end{itemize}
In order to estimate these quantities, we have to compute
averages of various functions with respect to the densities $\pi$ and $\hat{\rho}$.
In order to approximate these averages, we employ Markov Chain Monte Carlo methods, i.e. we use the Metropolis-Hasting algorithm to sample $\{\theta_{\pi,i}\}_{i=1}^{N_f}$, s.t. $\theta_{\pi,i}\sim \pi(\theta)$ and  $\{\theta_{\hat{\rho},i}\}_{i=1}^{N_f}$, s.t. $\theta_{\hat{\rho},i}\sim \hat{\rho}(\theta)$. The corresponding averages for $\pi$ and $\hat{\rho}$ 
are then approximated by computing the 
arithmetic means of the corresponding functions evaluated for $\{\theta_{\pi,i}\}_{i=1}^{N_f}$ and
respectively  for $\{\theta_{\hat{\rho},i}\}_{i=1}^{N_f}$.
Below we describe the details behind computing the various quantities mentioned above. 

\subsection*{Computing $\Phi(\pi,r)$ }
In theorem \ref{thm:pac:alt1} we are provided with an expression for $\Phi(\pi,r)$. 
For the purposes of the numerical example, we use the case when $r=2$.
For this case, the expression for $\Phi(\pi,r)$ is as follows:
\[\Phi(\pi,2)=3\mu_{max}(Q_e)\sqrt{(m+1)!}\sqrt{E_{\theta\sim \pi} \left (\frac{\hat{\rho}(\theta)}{\pi(\theta)} \right )^2}\sqrt{E_{\theta\sim \pi} G_e^4(\theta)}. \]

\begin{figure}
	\centering
	\includegraphics[width=0.7\linewidth]{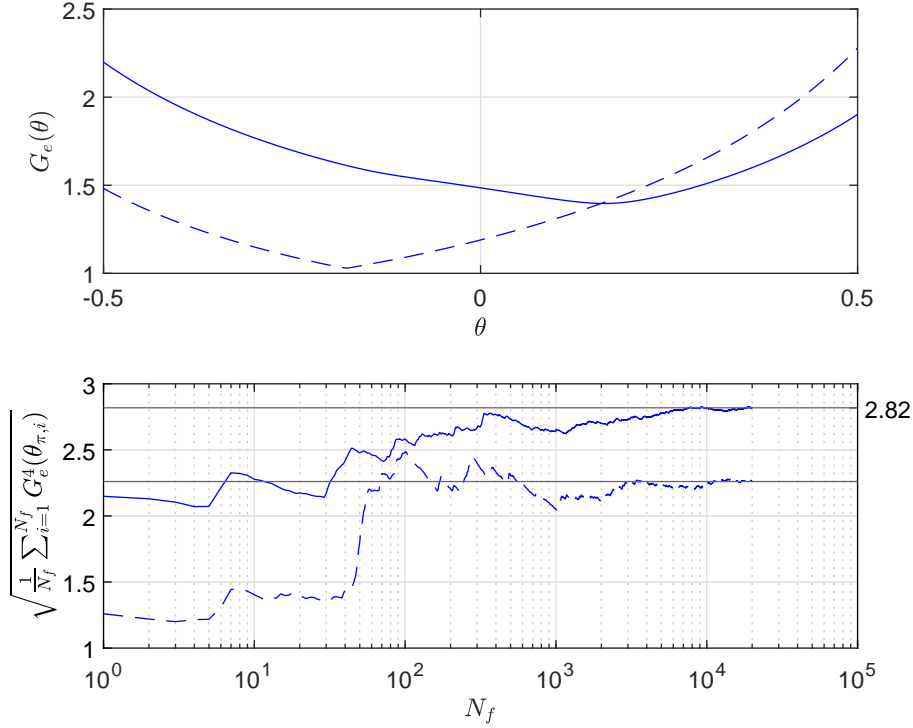}
	\caption{TOP: $G_e$ as function of $\theta$, Bottom: convergence of the approximation, solid lines show results for case 1: $\w=\mathbf{u}$, dashed lines show case 2: $\w=[\y^T,\mathbf{u}^T]^T$}
	\label{fig:Ge}
\end{figure}

In order to compute $E_{\theta\sim \pi} G_e^4(\theta)$, we use 
\begin{equation}
	\sqrt{E_{\theta\sim \pi} G_e^4(\theta)}\approx\sqrt{\frac{1}{N_f}\sum_{i=1}^{N_f}G_e^4(\theta_{\pi,i})}.\label{eq:Ge4approx}
\end{equation}
Given numerical value of $\theta_{\pi,i}$, one can construct a state-space system from \eqref{numEx:params}, and compute the sum defining $G_e$ (See Definition \ref{def:constants}) with finite number of terms until convergence of the sum, to obtain $G_e(\theta_{pi,i})$. 
In figure \ref{fig:Ge}, we see how $G_e(\theta)$ looks for this specific generating system and the two different hypothesis classes. Note, that for case 1 it achieves a minimum at the optimal $\theta=0.16$ parameter, $G_e$ for case 2 achieves minimum at $\theta=-0.17$. The bottom subfigure of figure \ref{fig:Ge} showcases that the monte carlo approximation of $\sqrt{E_{\theta\sim \pi} G_e^4(\theta)}$, does converge, and the horizontal line at $2.82$, denotes the monte carlo approximation with maximal number of samples tested, for case 1. 

Moving on to $\sqrt{E_{\theta\sim \pi} \left (\frac{\hat{\rho}(\theta)}{\pi(\theta)} \right )^2}$, normally one would need a normalization constant of $\pi(\theta)$, however because we use Gibbs posterior \eqref{eq:OLDgibbs}, we can simplify it to
\begin{align}
	\sqrt{E_{\theta\sim \pi} \left (\frac{\hat{\rho}(\theta)}{\pi(\theta)} \right )^2}=\sqrt{E_{\theta\sim \pi} \left (Z^{-1}e^{-\lambda\hat{\mathcal{L}}_N(\theta)} \right )^2}\approx \frac{Z^{-1}}{\sqrt{N_f}}\sqrt{\sum_{i=1}^{N_f} e^{-2\lambda\hat{\mathcal{L}}_N(\theta_{\pi,i})}} 
\end{align}
therefore we only need to estimate the normalisation constant of the posterior distribution $Z$
\begin{align}
	Z=E_{\theta\sim\pi}e^{-\lambda\hat{\mathcal{L}}_N(\theta)}\approx \frac{1}{N_f}\sum_{i=1}^{N_f}e^{-\lambda\hat{\mathcal{L}}_N(\theta_{\pi,i})}\\
	\ln(Z)\approx \ln\left (\sum_{i=1}^{N_f}e^{-\lambda\hat{\mathcal{L}}_N(\theta_{\pi,i})}\right )-\ln(N_f)
\end{align}

We will be using the log-sum-exponential trick to help with numerical computation, i.e.
$$lse(x)=lse(\begin{bmatrix} x_1&\dots&x_n\end{bmatrix})\triangleq\ln\left(\sum_i^n e^{x_i}\right)=max(x)+\ln\left(\sum_i^ne^{x_i-max(x)}\right) $$

\begin{align}
	\ln (Z)\approx lse(-\lambda\hat{\mathcal{L}}_N(\theta_{\pi}))-\ln (N_f)
\end{align}
where $\hat{\mathcal{L}}_N(\theta_{\pi})=[\hat{\mathcal{L}}_N(\theta_{\pi,1}),\dots \hat{\mathcal{L}}_N(\theta_{\pi,N_f})]$, then
\begin{align}
	\sqrt{E_{\theta\sim \pi} \left (\frac{\hat{\rho}(\theta)}{\pi(\theta)} \right )^2}\approx \exp\left (0.5lse(-2\lambda\hat{\mathcal{L}}_N(\theta_{\pi})) -\ln (Z)-0.5 \ln(N_f) \right )\\
	=\exp\left (0.5lse(-2\lambda\hat{\mathcal{L}}_N(\theta_{\pi})) -lse(-\lambda\hat{\mathcal{L}}_N(\theta_{\pi}))+0.5\ln (N_f) \right ) \label{eq:Drapprox}
\end{align}
 
\begin{figure}
	\centering
	\includegraphics[width=0.6\linewidth]{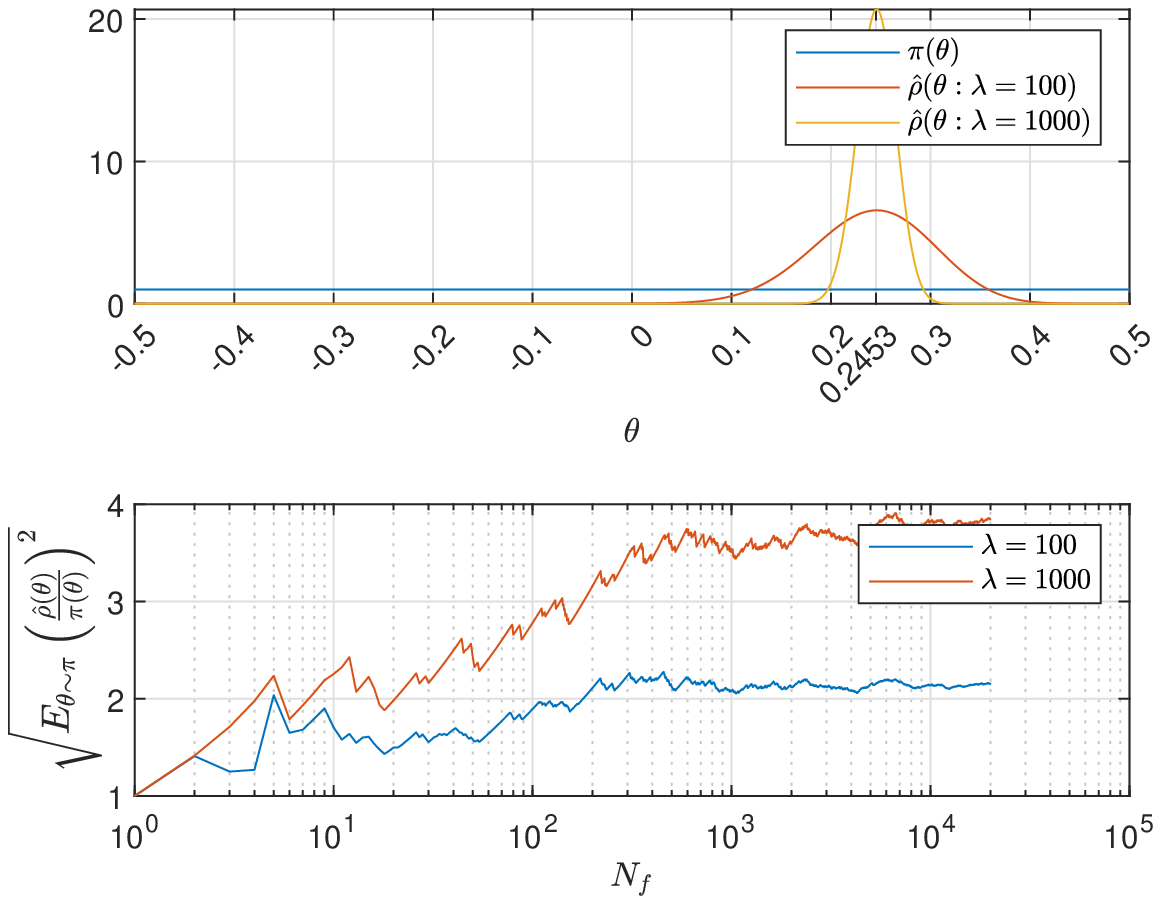}
	\caption{Case 1. Top: Prior and posterior candidates, for $N=100$, and several choices of $\lambda$. Bottom: Convergence of numerical approximation}
\end{figure}
\begin{figure}
	\centering
	\includegraphics[width=0.6\linewidth]{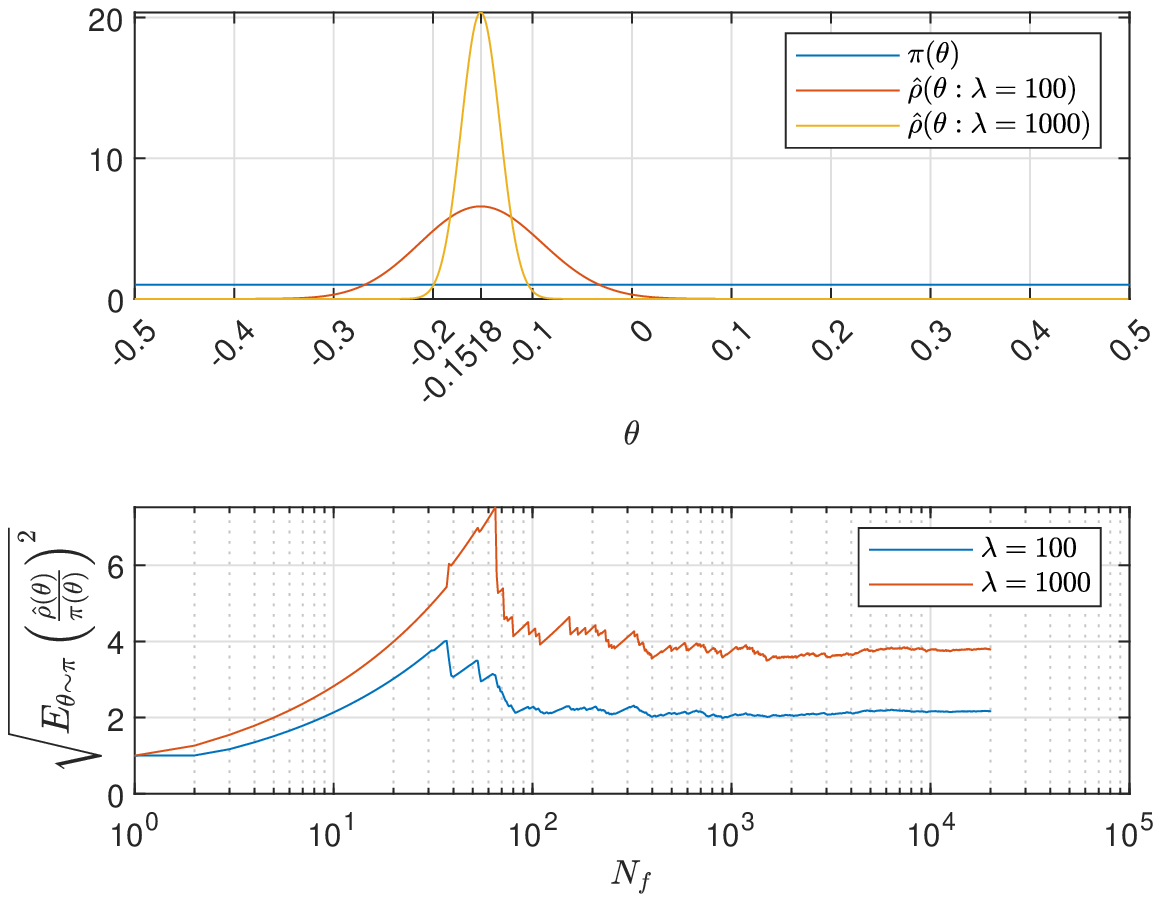}
	\caption{Case 2. Top: Prior and posterior candidates, for $N=100$, and several choices of $\lambda$. Bottom: Convergence of numerical approximation}
\end{figure}

\begin{figure}
	\centering
	\includegraphics[width=\linewidth]{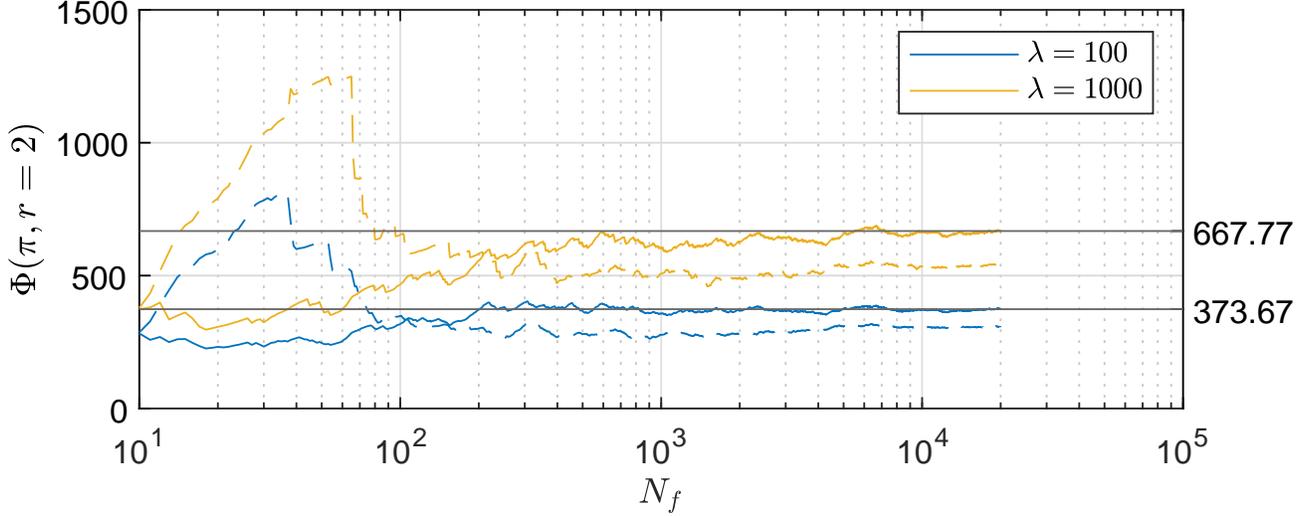}
	\caption{Convergence of $\Phi(\pi,r=2)$ for this specific system, hypothesis class $\mathcal{F}$, using Gibbs posterior, and $N=100$, solid lines show results for case 1: $\w=\mathbf{u}$, dashed lines show case 2: $\w=[\y^T,\mathbf{u}^T]^T$}
	\label{fig:Phi}
\end{figure}
Similarly to bottom subfigure of figure \ref{fig:Ge}, in figure \ref{fig:Phi}, we can see how $\Phi(\pi,2)$ converges with different number of monte carlo samples, and the horizontal lines denoting the approximation with maximum samples tested.\\

\subsection*{Computing $\frac{2}{\delta N} E_{\theta\sim \hat{\rho}} G(\theta)$}
We  approximate
\begin{align}
    E_{\theta\sim \hat{\rho}} G(\theta) \approx \frac{1}{N_f}\sum_{i=1}^{N_f} G(f_{\theta_i}) \label{eq:Gapprox}
\end{align}
For a given predictor $\Sigma(\theta_i)=\left (\hat{A}_i,\hat{B}_i,\hat{C}_i,\hat{D}_i \right )$ computing $G(\theta_i)$ is straightforward:  we have to compute
$G_0(f)$, $G_{-1}(f)$, $G_1(t)$, $G_2(f)$ for $f=f_{\Sigma(\theta_i)}$ and  $G_3(\w)$.
Computing $G_0(f)$, $G_{-1}(f)$, $G_1(f)$, $G_2(f)$ involves computing powers of matrices 
of $\Sigma(\theta_i)$ and the error system $(A_e,K_e,C_e)$ defined in Definition \ref{def:constants}.
Subsequently, infinite sums of the norms of these matrix powers have to be computed. 
The infinite sum  is approximated by adding summands until  the extra summand changes the total sum less than some tolerance. 
In order to compute $G_3(\w)$, we need the term $K_w$, which can be computed from the covariance of $\w$. In our case $K_w=4.18$ for the first case, and for the second case $K_w=4.62$.
\begin{figure}[H]
    \centering
    \includegraphics[width=0.7\linewidth]{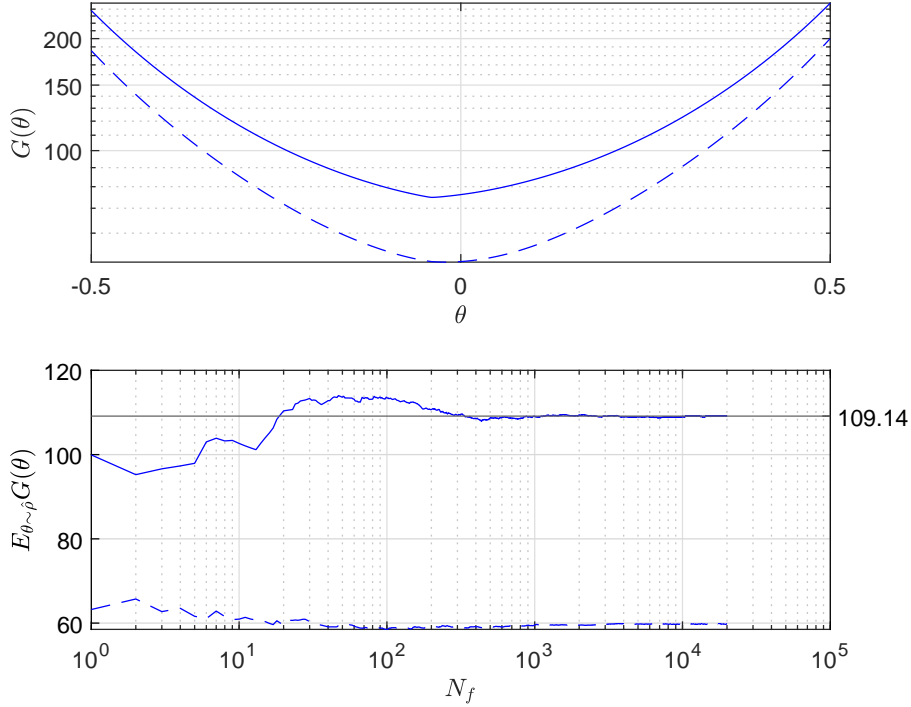}
    \caption{Numerical approximation of $G(\theta)$, solid lines show results for case 1: $\w=\mathbf{u}$, dashed lines show case 2: $\w=[\y^T,\mathbf{u}^T]^T$}
    \label{fig:Gs}
\end{figure}
In figure \ref{fig:Gs} we can see that $G(\theta)$ is an increasing function of $|\theta|$, and in this specific scenario, when minimising the PAC-Bayesian-Like upper bound w.r.t. $\rho$, $G(\theta)$ would act as regularisation term, minimising the absolute value of $\theta$.\\
Finally, taking approximation of $E_{\theta\sim\hat{\rho}} G(\theta)$ from \eqref{eq:Gapprox}, and $\Phi(\pi,2)$ approximation, we can compute the R\'enyi divergence based PAC-Bayesian-Like bound (theorem \ref{thm:pac:alt1}) 

\subsection*{Computing $KL(\hat{\rho} \|\pi)$ term}
In this section we will compute the Kullback–Leibler divergence. We can approximate the KL divergence using Monte Carlo method, by taking
\begin{align}
    KL(\hat{\rho} \|\pi)=E_{f\sim\hat{\rho}} \ln \frac{\hat{\rho}(f)}{\pi(f)} \approx \frac{1}{N_f}\sum_{i=1}^{N_f} \ln \frac{\hat{\rho}(\theta_{\rho,i})}{\pi(\theta_{\rho,i})}
\end{align}
However computing $KL$ in such a way requires $N_f$ to be quite large, so for this simple example where $\theta$ is scalar, we can compute $KL$ by approximating the integral
\begin{align}
    KL(\rho||\pi)&=\int_{-0.5}^{0.5}\rho(\theta) \ln \left (\frac{\rho(\theta)}{\pi(\theta)} \right )d\theta \approx \frac{1}{N_f}\sum_{i=1}^{N_f-1} \rho(\theta_i) \ln \left (\frac{\rho(\theta_i)}{\pi(\theta_i)} \right), \label{eq:KLapprox}
\end{align}
where $\theta_i$ are equally spaced from $0$ to $0.4$.

\subsection*{Computing $\widehat{\Psi}_{\pi}(\lambda,N)$}
Approximating the upper bound on moment generating function $\widehat{\Psi}_{\pi}(\lambda,N)$, involves computing $G_e(\Theta)=\sup_{\theta\in\Theta} G_e(\theta)$. We take advantage that we have samples $\theta_{\pi,i}$, which come from an uniform distribution, and as such
\begin{equation}
    G_e(\Theta)\approx \max(G_e(\theta_{\pi,i})) \label{eq:GeThetaapprox}
\end{equation}
Using $G_e(\Theta)$ approximation \eqref{eq:GeThetaapprox}, computing $\widehat{\Psi}_{\pi}(\lambda,N)$ 
involves simple arithmetic.

\subsection*{Computing generalised loss}
To compute the generalisation loss $\mathcal{L}(\theta)$ for some $f$, first compute the error system $f_e=f_{A_e,K_e,C_e,D_e}$, described in definition \ref{def:constants} and solve the Sylvester equation
\begin{align*}
    P=A_ePA_e^T+K_eQ_eK_e^T
\end{align*}
The equation above is a standard one in control theory and there are standard algorithms for
solving it. Then we can compute the generalization loss for a given parameter value as
\begin{align*}
    \mathcal{L}(\theta)=C_ePC_e^T+D_eQ_eD_e^T.
\end{align*}
If we take the samples  $\{\theta_{\rho,i}\}_{i=1}^{N_f}$, s.t. $\theta_{\rho,i}\sim \hat{\rho}(\theta)$, then we can approximate the average generelization loss by 
\begin{align}
    E_{\theta\sim \hat{\rho}} \mathcal{L} (\theta) \approx \frac{1}{N_f} \sum_{i=1}^{N_f} \mathcal{L}(\theta_{\rho,i})
\end{align}


\section{Proofs}
In this section we provide the proofs of theorem \ref{thm:pac:altKL} and \ref{thm:pac:alt1} under the assumptions stated in the main text. To do so we first prove a series of lemmas. 
\newcommand{\Efrho}{E_{f\sim\rho}}
\newcommand{\Efpi}{E_{f\sim\pi}}


\begin{lemma} \label{lemma:etoz} For random variable $\e_g(t)\sim\mathcal{N}(0,Q_e)$, the following holds
	\begin{align*} 
		\bE[\|\e_g(t)\|_2^r]\leq \mu_{\max}(Q_e)^{\frac{r}{2}}\bE[\|\z(t)\|_2^r]\\
		\z(t)\sim \mathcal{N}(0,I),
	\end{align*}
	where $Q_e=\bE[\e_g(t)\e_g^T(t)]$, and $\mu_{\max}(Q_e)$ denotes the maximal eigen value of $Q_e$.
\end{lemma}
\begin{proof}[Proof of Lemma \ref{lemma:etoz}]
	First, note $\z(t)=Q_e^{-\frac{1}{2}}\e_g(t)$, and 
	\begin{align*}
		\|\e_g(t)\|_2^2=\e_g^T(t)\e_g(t)=\z^T(t)Q_e^{\frac{1}{2}}Q_e^{\frac{1}{2}}\z(t)=\z^T(t)Q_e\z(t)
	\end{align*}
	therefore
	\begin{align*}
		\|\e_g(t)\|_2^2 \leq \mu_{\max}(Q_e)\|\z(t)\|_2^2\\
		\|\e_g(t)\|_2^r\leq\mu_{\max}(Q_e)^{\frac{r}{2}}\|\z(t)\|_2^r\\
		\bE[\|\e_g(t)\|_2^r]\leq\mu_{\max}(Q_e)^{\frac{r}{2}}\bE[\|\z(t)\|_2^r]\\
	\end{align*}
	Finally, note that $\z(t) \sim  \mathcal{N}(0,I)$.
\end{proof}
\begin{lemma} \label{lemma:zMomentsSq}
	If $\z(t)\sim \mathcal{N}(0,I_m)$, then
	\begin{align*}
		\bE[\|\z(t)\|_2^r]^2\leq 4((m+r-1)!)
	\end{align*}
\end{lemma}

\begin{proof}[Proof of Lemma \ref{lemma:zMomentsSq}]
	First, notice that the distribution of $\|\z(t)\|_2=\sqrt{\sum_{i=1}^m \z_i^2(t)}$ is chi- distribution, as such
	\begin{align}
		\bE[\|\z(t)\|_2^r]=2^{\frac{r}{2}}\dfrac{\Gamma(\frac{m+r}{2})}{\Gamma(\frac{m}{2})}
	\end{align}
	We will use mathematical induction to prove the lemma.\\
	\textbf{For $r=0$}, lemma holds, since
	\begin{align}
		\bE[\|\z(t)\|_2^0]^2=\left (2^{\frac{0}{2}}\dfrac{\Gamma(\frac{m+0}{2})}{\Gamma(\frac{m}{2})}\right )^2=1\leq 4(m-1)! 		,\quad \forall m\in\mathbb{N}.
	\end{align}
	\textbf{for $r=1$}, lemma holds, as
	\begin{align*}
		\bE[\|\z(t)\|_2^1]=2^{\frac{1}{2}}\dfrac{\Gamma(\frac{m+1}{2})}{\Gamma(\frac{m}{2})}.
	\end{align*}
	Notice that, for scalar $\x\sim\mathcal{N}(0,1)$
	\begin{align*}
		\bE[|\x|^k]=2^{\frac{k}{2}}\dfrac{\Gamma(\frac{k+1}{2})}{\sqrt{\pi}}
	\end{align*}
	It is also known that 
	\begin{align*}
		\bE[|\x|^k]=\begin{cases} (k-1)!!\sqrt{\frac{2}{\pi}},& k\text{ odd}\\ (k-1)!!,& k\text{ even} \end{cases}
	\end{align*}
	therefore,
	\begin{align*}
		2^{\frac{k}{2}}\dfrac{\Gamma(\frac{k+1}{2})}{\sqrt{\pi}}=\begin{cases} (k-1)!!\sqrt{\frac{2}{\pi}},& k\text{ odd}\\ (k-1)!!,& k\text{ even} \end{cases}
	\end{align*}
	Applying this to $k=m$ and $k=m-1$, we obtain
	\begin{align*}
		2^{\frac{m}{2}}\dfrac{\Gamma(\frac{m+1}{2})}{\sqrt{\pi}}=\begin{cases} (m-1)!!\sqrt{\frac{2}{\pi}},& m\text{ odd}\\ (m-1)!!,& m\text{ even} \end{cases}\\
		2^{\frac{m-1}{2}}\dfrac{\Gamma(\frac{m}{2})}{\sqrt{\pi}}=\begin{cases} (m-2)!!\sqrt{\frac{2}{\pi}},& (m-1)\text{ odd}, (m \text{ even})\\ (m-2)!!,& (m-1)\text{ even}, (m \text{ odd}) \end{cases}
	\end{align*}
	Now notice,
	\begin{align*}
		\bE[\|z(t)\|_2^1]=2^{\frac{1}{2}}\dfrac{\Gamma(\frac{m+1}{2})}{\Gamma(\frac{m}{2})} = \frac{ 2^{\frac{m}{2}}\dfrac{\Gamma(\frac{m+1}{2})}{\sqrt{\pi}}}{2^{\frac{m-1}{2}}\dfrac{\Gamma(\frac{m}{2})}{\sqrt{\pi}}}=\frac{(m-1)!!}{(m-2)!!}c_m\\
		c_m=\begin{cases}\sqrt{\frac{2}{\pi}} ,& m\text{ even}\\\sqrt{\frac{\pi}{2}} ,& m\text{ odd} \end{cases}
	\end{align*}
	notice that $c_m\leq 2$ for all $m$, and therefore
	\begin{align}
		\bE[\|\z(t)\|_2^1]\leq 2\frac{(m-1)!!}{(m-2)!!}\leq 2(m-1)!!
	\end{align}
	Then
	\begin{align*}
		\bE[\|\z(t)\|_2^1]^2\leq 4((m-1)!!)^2
	\end{align*}
	Note that $((m-1)!!)^2\leq m!$. We can see that by contradiction: assume that
		$((m-1)!!)^2\geq m !$. Notice that $m!=m!!(m-1)!!$ and hence $((m-1)!!)^2\geq m !$
		implies 
		$(m-1)!!\geq m!!$.
	As $(m-1)!!$ must be less than $m!!$ we have a contradiction. Therefore $((m-1)!!)^2\leq m!$ holds and we have
	\begin{align*}
		\bE[\|\z(t)\|_2^1]^2\leq 4 m!.
	\end{align*}
	That is, we have shown that for $r=0$ and $r=1$ Lemma \ref{lemma:zMomentsSq} holds. \\
	 Now suppose that for all $k \geq 2$ and for all  $0\leq r \leq k$
	\begin{align}
		2^{\frac{r}{2}}\dfrac{\Gamma(\frac{m+r}{2})}{\Gamma(\frac{m}{2})}\leq 4(m+r-1)!,
		\label{pf:b2:eq-1}
	\end{align}
	We will show that \eqref{pf:b2:eq-1} holds for $r=k+1$ too.
	To this end, notice that 
	\begin{align*}
		\Gamma \left (\frac{m+k}{2} \right )=\Gamma \left(\frac{m+k-2}{2}+1 \right)=\frac{m+k-2}{2}\Gamma \left(\frac{m+k-2}{2} \right)
	\end{align*}
	Using this relation we obtain
	\begin{equation}
	\label{pf:b2:eq1}
		\begin{split}
			\left (2^{\frac{k}{2}}\dfrac{\Gamma(\frac{m+k}{2})}{\Gamma(\frac{m}{2})}\right )^2=\left ( \left ( 2^{\frac{k-2}{2}}\dfrac{\Gamma(\frac{m+k-2}{2})}{\Gamma(\frac{m}{2})} \right )\left (2\frac{m+k-2}{2} \right) \right )^2 \\
			=\left ( 2^{\frac{k-2}{2}}\dfrac{\Gamma(\frac{m+k-2}{2})}{\Gamma(\frac{m}{2})} \right )^2\left (2\frac{m+k-2}{2}  \right )^2.
		\end{split}
	\end{equation}	
	Now $k-2\in[0,k]$, so we can apply to it the induction hypothesis. That is,
	for $r=k-2$, \eqref{pf:b2:eq-1} holds, i.e., 
	\begin{align*}
		\left ( 2^{\frac{r}{2}}\dfrac{\Gamma(\frac{m+r}{2})}{\Gamma(\frac{m}{2})} \right )\leq 4(m+r-1)!=4(m+k-3)!.
	\end{align*}
	and therefore
	\begin{align*}
		\left (2^{\frac{k}{2}}\dfrac{\Gamma(\frac{m+k}{2})}{\Gamma(\frac{m}{2})}\right )^2\leq 4(m+k-3)! \left ( 4\frac{(m+k-2)^2}{4} \right )\\
		=4(m+k-3)!(m+k-2)(m+k-2).
	\end{align*}
    Using $(m+k-2)\leq (m+k-1)$, it follows that 
	\begin{align*}
	 \left ( 2^{\frac{k-2}{2}}\dfrac{\Gamma(\frac{m+k-2}{2})}{\Gamma(\frac{m}{2})} \right )^2\left (2\frac{m+k-2}{2}  \right )^2 \le 
		4(m+k-3)!(m+k-2)(m+k-2)	\leq 4(m+k-1)!
	\end{align*}
	 Substituting the last inequality into \eqref{pf:b2:eq1}, it follows that
	 \eqref{pf:b2:eq-1} holds for $r=k+1$. 
\end{proof}
\begin{lemma}\label{lemma:evenzmoments} For random variable $\z\sim\mathcal{N}(0,I_m)$, the even moments of $\|\z\|_2$ are bounded by
	\begin{align*}
		\bE[\|\z\|_2^{2r}]\leq 2^r(m+r-1)!
	\end{align*}
	\begin{proof}[Proof of Lemma \ref{lemma:evenzmoments}]
		Clearly $\|\z\|_2$ has the chi distribution,
		\begin{align*}
			\bE[\|\z\|_2^{2r}]=2^{\frac{2r}{2}}\dfrac{\Gamma(\frac{m+2r}{2})}{\Gamma(\frac{m}{2})}=2^{r}\dfrac{\Gamma(\frac{m}{2}+r)}{\Gamma(\frac{m}{2})}
		\end{align*}
		\begin{align*}
			\Gamma\left (\frac{m}{2}+r \right )=\Gamma \left (\frac{m}{2}+(r-1) +1 \right )= \left (\frac{m}{2}+(r-1) \right )\Gamma \left (\frac{m}{2}+(r-1) \right )\\
			=\left (\frac{m}{2}+(r-1) \right )\left (\frac{m}{2}+(r-2) \right ) \dots \frac{m}{2}\Gamma\left ( \frac{m}{2} \right )
		\end{align*}
		\begin{align*}
			\bE[\|\z\|_2^{2r}]=2^r \frac{\left (\frac{m}{2}+(r-1) \right )\left (\frac{m}{2}+(r-2) \right ) \dots \frac{m}{2}\Gamma\left ( \frac{m}{2} \right )}{\Gamma\left ( \frac{m}{2} \right )}
		\end{align*}
		notice $\frac{m}{2}\leq m$, then
		\begin{align*}
			\bE[\|\z\|_2^{2r}]\leq 2^r\frac{(m+r-1)!}{m!} \leq 2^r (m+r-1)!
		\end{align*}
	\end{proof}
\end{lemma}
Combining Lemmas (\ref{lemma:etoz} and \ref{lemma:zMomentsSq}), we obtain the following lemma.
\begin{lemma} \label{lemma:evenEmoments}
	\begin{align*}
		\bE[\|\e_g(t)\|_2^{2r}]\leq \mu_{\max}(Q_e)^{r}2^r(m+r-1)!
	\end{align*}
\end{lemma}
Combining Lemmas (\ref{lemma:etoz} and \ref{lemma:evenzmoments}), we obtain the following lemma.
\begin{lemma} \label{lemma:Emoments}
	\begin{align*}
		\bE[\|\e_g(t)\|_2^r]\leq 2\mu_{\max}(Q_e)^{\frac{r}{2}}\sqrt{(m+r-1)!}
	\end{align*}
\end{lemma}
\begin{lemma}\label{lemma:boundOnCovarianceW} The covariance $\Lambda_{k-j}^{\w}=\bE[\w(t-j)\w^T(t-k)]$ of the stationary process $\w$ is bounded in norm
	\begin{align}
		&\|\Lambda_{k-j}^{\w}\|_2\leq K_w\\
		&K_w=\max \{K_{w,1},K_{w,2}\}\nonumber \\
		&K_{w,1}= \left(\|C_w\|_2^2\|K_g\|_2^2+\|C_w\|_2 \|K_g\|_2 \right)\left (\sum_{k=0}^{\infty} \|A_g^{k}\|_2\right )^3\hspace{-5pt} \mu_{\max}(Q_e)\nonumber\\
		&K_{w,2}= \left( \|C_w\|_2^2\|K_g\|_2^2+1 \right) \left( \sum_{k=0}^{\infty} \|A_g^{k}\|_2 \right)^2\hspace{-5pt} \mu_{\max}(Q_e) \nonumber
	\end{align}
	Note that, in the case of $\w=[\y^T,\mathbf{u}^T]^T$ the matrix $C_w=C_g$.
\end{lemma}
\begin{proof}[Proof of Lemma \ref{lemma:boundOnCovarianceW}]
	{\small
		\begin{align*}\\
			\Lambda_{k-j}^{\w}=\begin{cases}C_wA_g^{k-j-1}\left(A_gP_gC_w^T+K_gQ_e\begin{bmatrix} 0\\I\end{bmatrix}\right), &k-j>0\\
				C_wP_gC_w^T+\begin{bmatrix}0&I\end{bmatrix}Q_e\begin{bmatrix}0\\I\end{bmatrix}, &k-j=0\\
				(\Lambda_{j-k}^{\w})^T, &k-j<0\end{cases}
		\end{align*}
	}
	\begin{align}
	     P_g = \sum_{k=0}^{\infty}A_g^kK_gQ_eK_g^T(A_g^k)^T \label{eq:boundNormPg1} \\
		\| P_g\|_2 &=\| \sum_{k=0}^{\infty}A_g^kK_gQ_eK_g^T(A_g^k)^T \|_2  \label{eq:boundNormPg}\\
		& \le \|K_g\|_2^2\mu_{\max}(Q_e)\sum_{k=0}^{\infty} \|A_g^{k}\|_2^2 \nonumber \\
		& \le \|K_g\|_2^2\mu_{\max}(Q_e)\left(\sum_{k=0}^{\infty} \|A_g^{k}\|_2\right)^2 \nonumber 
	\end{align}
	\begin{align}
		&\| \Lambda_{k-j>0}^{\w} \|_2 \leq \|C_w\|_2^2\|A_g^r\|_2 \|P_g\|_2 +\|C_w\|_2^ \|A_g^{r-1}\|_2\|K_g\|_2)\mu_{\max}(Q_e) \nonumber \\
		&\leq \|C_w\|_2^2\|A_g^r\|_2\|K_g\|_2^2\mu_{\max}(Q_e) \left(\sum_{k=0}^\infty \|A_g^{k}\|_2\right)^2 +  \|C_w\|_2 \|K_g\|_2 \sum_{k=0}^\infty \|A_g^{k}\|_2 \mu_{\max}(Q_e) \nonumber\\
		&\leq \left (\|C_w\|_2^2 \left (\sum_{k=0}^\infty \|A_g^{k}\|_2\right) \|K_g\|_2^2 \left(\sum_{k=0}^\infty \|A_g^{k}\|_2\right)^2+ \|C_w\|_2 \|K_g\|_2 \sum_{k=0}^\infty \|A_g^{k}\|_2\right)  \mu_{\max}(Q_e) \nonumber 
	\end{align}
	Since $\sum_{k=0}^{\infty} \|A_g^{k}\| \ge 1$, it follows that
	$\sum_{k=0}^{\infty} \|A_g^{k}\| \le (\sum_{k=0}^{\infty} \|A_g^{k}\|)^3$
	therefore
	\begin{align*}
		&\| \Lambda_{k-j>0}^{\w} \|_2 \leq (\|C_w\|_2^2\|K_g\|_2^2+\|C_w\|_2 |K_g\|_2)\left(\sum_{k=0}^{\infty} \|A_g^{k}\|_2\right)^3  \mu_{\max}(Q_e)
	\end{align*}
	Similarly
	\begin{align*}
		\| \Lambda_{k-j=0}^{\w} \|_2 \leq (\|C_w\|_2^2\|K_g\|_2^2+1) \left(\sum_{k=0}^{\infty} \|A_g^{k}\|_2\right)^2 \mu_{\max}(Q_e)
	\end{align*}
\end{proof}
\begin{lemma}\label{lem:Ev-hl} With notation as stated above, the following holds true
    \begin{equation}
        \bE[|V_N(f)-\hat{\mathcal{L}}_N(f)|]\leq \frac{2}{N}G(f)
    \end{equation}
    where $G(f)=G_{-1}(f)G_0(f)G_1(f)G_2(f)G_3(\w)$
    \begin{align*}
        G_{-1}(f)&\sqrt{2 \left(\left(\sum_{k=0}^{\infty}  \|A_e^k \|_2^2 \right) +2 \right)},\\
        G_0(f)&=   \sum_{k=0}^{\infty} \| \hat{A}^k \|_2,\\
        G_1(f)&=\|(A_e,K_e,C_e,D_e)\|_{H_2,*} \triangleq \sqrt{\|D_e\|_2^2 +\sum_{k=0}^{\infty} \|C_e\|_2^2\|A_e^k\|_2^{2}\|B_e\|_2^2}, \\
        G_2(f)&=\|(\hat{A},\hat{B},\hat{C},\hat{D})\|_{\ell_1,*} \triangleq\|\hat{D}\|_2 +\sum_{k=0}^{\infty} \|\hat{C}\|_2\|\hat{A}^k\|_2\|\hat{B}\|_2,\\
		G_3(\w)&= \sqrt{\mu_{\max}(Q_e)K_w},\\
		\|\bE[\w(t)\w^T(t-r)]\|_2&\leq K_w
    \end{align*}
\end{lemma}
\begin{proof}[Proof of Lemma \ref{lem:Ev-hl}] 
	{\newcommand{\trace}[1]{\text{trace}\left ( #1 \right )}
		\newcommand{\hzf}{\hat{\mathbf{z}}_f}
		\begin{align*}
			&\bE[|V_N(f)-\hat{\mathcal{L}}_N(f)|]\leq\frac{1}{N}\sum_{t=0}^{N-1}\bE[|(\underset{\hzf(t)}{\underbrace{\y(t)-\hyfi(t)}})^2-(\underset{\hzf(t\mid 0)}{\underbrace{\y(t)-\hyfi(t|0)}})^2 |]
		\end{align*}
		For ease of notation let us define
		\begin{align*}
			\hzf(t)=\y(t)-\hyfi(t)\\
			\hzf(t\mid 0)=\y(t)-\hyfi(t\mid 0)
		\end{align*}
		then
		\begin{align*}
			&\bE[|V_N(f)-\hat{\mathcal{L}}(f)|]\\
			&\quad \leq \frac{1}{N}\sum_{t=0}^{N-1}\bE[|(\hzf(t)-\hzf(t\mid 0))(\hzf(t)+\hzf(t\mid 0)|]\\
			&\quad \leq\frac{1}{N}\sum_{t=0}^{N-1}\bE[|(\hzf(t)-\hzf(t\mid 0))||(\hzf(t)+\hzf(t\mid 0))|]\\
			&\quad \leq\frac{1}{N}\sum_{t=0}^{N-1}\sqrt{\bE[(\hzf(t)-\hzf(t\mid 0) )^2]} \sqrt{\bE[(\hzf(t)+\hzf(t\mid 0))^2]}
		\end{align*}
		We will separately look into the two terms $\sqrt{\bE[(\hzf(t)-\hzf(t\mid 0) )^2]}$ and $\sqrt{\bE[(\hzf(t)+\hzf(t\mid 0))^2]}$
		\begin{align*}
			&\bE[|(\hzf(t)-\hzf(t\mid0))^2|]=\bE[(\hyfi(t\mid 0)-\hyfi(t))^2]=\bE[(\hyfi(t)-\hyfi(t\mid 0))^2]
		\end{align*}
		\begin{align*}
			\hyfi(t)-\hyfi(t\mid 0)&=\left ( \sum_{k=0}^\infty \hat{C}\hat{A}^{k-1}\hat{B}\w(t-k)+\hat{D}\w(t)\right )-\left ( \sum_{k=0}^{t} \hat{C}\hat{A}^{k-1}\hat{B}\w(t-k)+\hat{D}\w(t)\right )\\
			& =\sum_{k=t+1}^\infty \hat{C}\hat{A}^{k-1}\hat{B}\w(t-k)
		\end{align*}
		As process $\w$ is stationary, let us define lag dependant covariance as
		\begin{align*}
			\Lambda_{k-j}^{\w}=\bE[\w(t)\w^T(t-(k-j))]=\bE[\w(t-k)\w^T(t-j)]
		\end{align*}
		and from this it follows
		\begin{align*}
			&\bE[(\hyfi(t)-\hyfi(t\mid 0))^2]=\sum_{k=t+1}^\infty\sum_{j=t+1}^\infty  \text{trace}\left ( \hat{C}\hat{A}^{k-1}\hat{B}\Lambda_{k-j}^{\w}\hat{B}^T(\hat{A}^{j-1})^T\hat{C}^T \right )
		\end{align*}
		As we have assumed that $\y(t)$ is scalar, then we can drop the trace.
		Using Lemma \ref{lemma:boundOnCovarianceW}, for $\|\Lambda_{k-j}^{\w}\|\leq K_w$, we have
		\begin{align*}
			&\bE[(\hyfi(t)-\hyfi(t\mid 0))^2]\\
			&\leq \|\hat{C}\|_2^2\|\hat{B}\|_2^2K_w\sum_{k=t+1}^\infty\|\hat{A}^{k-1}\|_2\sum_{j=t+1}^\infty  \|\hat{A}^{j-1} \|_2 \\
			& = \|\hat{C}\|_2^2\|\hat{B}\|_2^2K_w (\sum_{k=t+1}^\infty\|\hat{A}^{k-1}\|_2)^2 \\
			& \le \|\hat{C}\|_2^2\|\hat{B}\|_2^2K_w \|\hat{A}^{t}\|_2^2 (\sum_{k=0}^{\infty} \|\hat{A}^k\|_2)^2 
		\end{align*}
		In the last step we used the fact that 
		$\| \hat{A}^k \|_2 = \| \hat{A}^t \hat{A}^{k-t} \|_2 \le \|\hat{A}^t\|_2 \|\hat{A}^{k-t} \|_2$
		and hence 
		\begin{align*}
			\sum_{k=t+1}^\infty\|\hat{A}^{k-1}\|_2  \le \|\hat{A}^{t}\|_2 \sum_{k=0}^{\infty} \|\hat{A}^k\|_2
		\end{align*}
		
		We now have the first term necessary
		\begin{align}
			&\sqrt{\bE[(\hzf(t)-\hzf(t\mid0))^2]}\leq \|\hat{C}\|_2\|\hat{B}\|_2\sqrt{K_w} 
			\|\hat{A}^t\|_2 \sum_{k=0}^{\infty} \|\hat{A}^k\|_2
		\end{align}
		Now for the second term
		\begin{align*}
			\bE[(\hzf(t)+\hzf(t\mid 0))^2]=\bE[\hzf^2(t)]+\bE[\hzf^2(t\mid 0)]+2\bE[\hzf(t)\hzf(t\mid 0)]
		\end{align*}
		which can be bounded using arithmetic mean by
		\begin{align*}
			\bE[(\hzf(t)+\hzf(t\mid 0))^2]\leq 2(\bE[\hzf^2(t)]+\bE[\hzf^2(t\mid 0)])
		\end{align*}
		Now let's focus on $\bE[\hzf^2(t)]$. First, notice that if
		$X$ is a vector, then $X^TQ_eX \le \mu_{max}(Q_e)\|X\|_2^2$. 
		Applying this remark to $X=(C_eA_e^kK_e)^T$ (note that $C_e$ has one row)  and to
		$X=D_e^T$ and 
		noticing that the euclidian norm of
		a vector and its induced norm as a matrix coincide, 
		\begin{equation}
		\label{pf:b7:eq0}
		\begin{split}
		& \underbrace{C_eA_e^kK_e}_{X^T} Q_e \underbrace{(C_eA_e^kK_e)^T}_{X}
		 \le \mu_{max}(Q_e) \|C_eA_e^kK_e\|_2^2 \le \|C_e\|_2^2 \|A_e^k\|^2_2 \|K_e\|_2^2 \\
		& \underbrace{D_e}_{X^T}Q_e\underbrace{D_e^T}_{X} \le \mu_{max}(Q_e) \|D_e\|_2^2 
		\end{split} 
		\end{equation}
		Using \eqref{pf:b7:eq0}, the following can be derived: 
		\begin{align*}
			\bE[\hzf^2(t)]&=C_{e}P_eC_{e}^T+D_{e}Q_eD_{e}\\
			&=\sum_{k=0}^\infty C_{e}A_e^kK_eQ_eK_e^T(A_e^k)^TC_{e}^T+D_{e}Q_eD_{e}\\
			\bE[\hzf^2(t)]&\leq \left ( \sum_{k=0}^{\infty} \|C_{e}\|_2^2\|A_e^k\|_2^2 \|K_e\|_2^2+\|D_{e}\|_2^2 \right )\mu_{\max}(Q_e) 
		\end{align*}
		defining $\|(A,K,C,D)\|_{H_2,*}\triangleq\sqrt{\|D\|_2^2 +\sum_{k=0}^{\infty} \|C\|_2^2\|A^k\|_2^{2}\|B\|_2^2}$ Note that 
		$\|(A,K,C,D)\|_{H_2,*}$
		 can be thought as an upper bound on the well-known $H_2$ norm of LTI systems \cite{HanzonBook,Katayama:05}, i.e. $\|(A,B,C,D)\|_{H_2} \le \|(A,B,C,D)\|_{H_2,*}$. When applied to the error system $(A_e,K_e,C_e,D_e)$ we will denote it as $G_1(f)$ With this definition we obtain
		\begin{align}
		    \bE[\hzf^2(t)]&\leq G_1(f)^2 \mu_{\max}(Q_e).
		\end{align}
		For $\bE[\hzf^2(t\mid 0)]$ we can again use \eqref{pf:b7:eq0},
		\begin{align*}
			& \bE[\hzf^2(t\mid 0)]=C_{e}P_{e,t}C_{e}^T+D_{e}Q_eD_{e}^T\\
			& =C_{e}A_e^tP_0(A_e^t)^TC_{e}^T+\sum_{k=0}^{t-1}C_{e}A_e^kK_eQ_eK_e^T(A_e^k)^TC_{e}^T +    D_{e}Q_eD_{e}^T\\
			&\leq C_{e}A_e^tP_0 (A_e^t)^TC_e^T +\|C_{e}\|_2^2\|K_e\|_2^2\mu_{\max}(Q_e)\sum_{k=0}^{t-1}\|A_e^{k}\|_2^2 +\|D_{e}\|_2^2\mu_{\max}(Q_e)\\
			&\leq C_{e}A_e^{t}P_0(A_e^t)^TC_e^T + (\sum_{k=0}^{\infty} \|C_{e}\|_2^2\|A_e^{k}\|_2^2 \|K_e\|_2^2+\|D_{e}\|_2^2)\mu_{\max}(Q_e) 
		\end{align*}
		Now $P_0=\begin{bmatrix} P_g & 0\\ 0 & 0 \end{bmatrix}$, as such
		$P_0=\begin{bmatrix} I \\ 0 \end{bmatrix}P_g \begin{bmatrix} I \\ 0 \end{bmatrix}^T$.
		Using  \eqref{eq:boundNormPg1},
		\begin{equation}
		\label{pf:b7:eq1}
			C_eA_e^t P_0 (A_e^t)^TC_e^T = 
			   \sum_{k=0}^{\infty} C_e A_e^{t}\begin{bmatrix} I \\ 0 \end{bmatrix}  A_g^kK_gQ_eK_g^T(A_g^k)^T(A_e^{t}\begin{bmatrix} I \\ 0 \end{bmatrix})^TC_e^T
		\end{equation}	
		Notice that $X=(C_e A_e^{t}\begin{bmatrix} I \\ 0 \end{bmatrix}  A_g^kK_g)^T$ is a 
		vector, hence, 
		\begin{equation}
		 \label{pf:b7:eq2} 
		 \begin{split}
		 & \underbrace{C_e A_e^{t}\begin{bmatrix} I \\ 0 \end{bmatrix}  A_g^kK_g}_{X^T} Q_e\underbrace{K_g^T(A_g^k)^T(A_e^{t}\begin{bmatrix} I \\ 0 \end{bmatrix})^TC_e^T}_{X} \le \\
		 & \mu_{\max}(Q_e)  \| C_e A_e^{t} \begin{bmatrix} I \\ 0 \end{bmatrix}  A_g^kK_g\|^2_2 \le \\
		  & \mu_{\max}(Q_e)  \| C_e\|_2^2 \|K_g\|_2^2 \|A_e^{t} \begin{bmatrix} I \\ 0 \end{bmatrix}  A_g^k\|^2_2
		  \end{split}	
		  \end{equation}
		  Using \eqref{pf:b7:eq1} and \eqref{pf:b7:eq2}, we can derive the following
		\begin{align*}
		  	C_eA_e^t P_0 (A_e^t)^TC_e^T =
			   \sum_{k=0}^{\infty} \|C_eA_e^{t}\begin{bmatrix} I \\ 0 \end{bmatrix}  A_g^kK_gQ_eK_g^T(A_g^k)^T(A_e^{t}\begin{bmatrix} I \\ 0 \end{bmatrix})^T C_e^T \|_2 \\
			 \leq \|C_e\|_2^2 \|K_g\|_2^2\mu_{\max}(Q_e)  \left( \sum_{k=0}^{\infty} \| A_e^{t}\begin{bmatrix} I \\ 0 \end{bmatrix} A_g^k \|_2^2 \right) \\
			=\|C_e\|_2^2 \|K_g\|_2^2\mu_{\max}(Q_e)  \left( \sum_{k=0}^{\infty} \| A_e^{t}\|_2^2 \|A_g^k \|_2^2 \right) \\
			=\|C_e\|^2_2 \|K_g\|_2^2\mu_{\max}(Q_e) \| A_e^{t}\|^2_2 \left( \sum_{k=0}^{\infty}  \|A_g^k \|_2^2 \right)
		\end{align*}
		Notice that for the case where $\w=\mathbf{u}$
		\begin{align}
			\|K_e\|_2^2&=\sup_{\|v\|_2=1}\left\|\begin{bmatrix} K_g\\ \begin{bmatrix} 0 & \hat{B} \end{bmatrix} \end{bmatrix} v \right\|_2^2 \nonumber\\
			&=\sup_{\|v\|_2=1} \|K_gv\|_2^2+\|\begin{bmatrix} 0 & \hat{B} \end{bmatrix}v\|_2^2 \geq \|K_g\|_2^2 \label{eq:KeBound}
		\end{align}
		In the second case where $\w=[\y^T,\mathbf{u}^T]^T$, instead of $\begin{bmatrix} 0 & \hat{B}\end{bmatrix}$ we have $\hat{B}'$, this does not change \eqref{eq:KeBound}, and $\|K_e\|_2^2\geq \|K_g\|_2^2$ in both cases.
		
		Then for both cases we have 
		\begin{align*}
			\|A_e^k \|^2_2&=\sup_{\|u\|_2^2+\|v\|_2^2=1}\left \|\begin{bmatrix} A_g^k & 0 \\ \star & \hat{A}^k \end{bmatrix} \begin{bmatrix} u \\ v \end{bmatrix} \right \|_2^2 \\
			&\ge \sup_{\|u\|_2^2=1} \left \|\begin{bmatrix} A_g^k & 0 \\ \star & \hat{A}^k \end{bmatrix} \begin{bmatrix} u \\ 0 \end{bmatrix} \right \|_2^2 \\
			&=\sup_{\|u\|_2=1} \|A_g^{k} u\|_2^2  + \| \star u\|^2_2   
			\ge\sup_{\|u\|_2=1} \|A_g^{k} u\|_2^2=\|A_g^k \|_2^2 
			\\
		\end{align*}
		Hence, by using $\| A_e^t \|_2^2 \le \sum_{k=0}^{\infty} \|A_e^k\|_2^2$
		and $\sum_{k=0}^{\infty}  \|A_g^k \|_2^2 \le \sum_{k=0}^{\infty} \|A_e^k\|_2^2$
		\begin{align*}
			\|K_g\|_2^2\mu_{\max}(Q_e) \| A_e^{t}\|_2 \left( \sum_{k=0}^{\infty}  \|A_g^k \|_2^2 \right) \le \|K_e\|_2^2\mu_{\max}(Q_e) \left( \sum_{k=0}^{\infty}  \|A_e^k \|_2^2 \right)^2
		\end{align*}
		Combining all this, it follows that 
		\begin{align*}
			\bE[\hzf^2(t\mid 0)] \leq \|C_{e}\|_2^2 
			\|K_e\|_2^2 \mu_{\max}(Q_e) \left( \sum_{k=0}^{\infty} \| A_e^k\|_2^2\right)^2 +G_1(f)^2 \mu_{\max}(Q_e) \\
			=\mu_{\max}(Q_e) \left [G_1(f)^2 \left(\sum_{k=0}^{\infty}  \|A_e^k \|_2^2 \right) + G_1(f)^2  \right ]\\
			=\mu_{\max}(Q_e) G_1(f)^2 \left(\left(\sum_{k=0}^{\infty}  \|A_e^k \|_2^2 \right) +1 \right)
		\end{align*}
		combining it together we get
		\begin{align*}
			&\bE[(\hzf(t)+\hzf(t\mid 0))^2]\leq 2(\bE[\hzf^2(t)]+\bE[\hzf^2(t\mid 0)])\\
			&\bE[(\hzf(t)+\hzf(t\mid 0))^2] \leq 2 G_1(f)^2 \mu_{\max}(Q_e)\left(\left(\sum_{k=0}^{\infty}  \|A_e^k \|_2^2 \right) +2 \right)
		\end{align*}
		For notation let us define $G_{-1}(f)\triangleq \sqrt{2 \left(\left(\sum_{k=0}^{\infty}  \|A_e^k \|_2^2 \right) +2 \right)}$, and so
		\begin{align*}
			\sqrt{\bE[|(\hzf(t)+\hzf(t\mid 0))^2|]}\leq \sqrt{\mu_{\max}(Q_e)}G_1(f)G_{-1}(f)
		\end{align*}
		Now to bring it all back together, recall
		\begin{align*}
			&\bE[|V_N(f)-\hat{\mathcal{L}}(f)|]\leq\frac{1}{N}\sum_{t=0}^{N-1}\sqrt{\bE[(\hzf(t)-\hzf(t\mid0))^2]}\sqrt{\bE[(\hzf(t)+\hzf(t\mid 0))^2]}\\
			&\leq\frac{2}{N}\sqrt{\mu_{\max}(Q_e)}G_1(f)G_{-1}(f) \|\hat{C}\|_2\|\hat{B}\|_2\sqrt{K_w} \left (\sum_{k=0}^{\infty} \|\hat{A}^k\|_2 \right ) \sum_{t=0}^{N-1} \|\hat{A}^t\|_2\\
			&\leq \frac{2}{N}\sqrt{\mu_{\max}(Q_e)}G_1(f)G_{-1}(f) \|\hat{C}\|_2\|\hat{B}\|_2\sqrt{K_w} \left (\sum_{k=0}^{\infty} \|\hat{A}^k\|_2 \right )^2 
		\end{align*}
		We define $\|(A,B,C,D)\|_{\ell_1,*} \triangleq \|D\|_2 +\sum_{k=0}^{\infty} \|C\|_2\|A^k\|_2\|B\|_2$, as an upper bound on the well-known $\ell_\infty$ norm of LTI systems \cite{DahlehTACFeedback}. When apllied to a predictor $(\hat{A},\hat{B},\hat{C},\hat{D})$, we will denote it as $G_2(f)$, i.e. $G_2(f)=\|(\hat{A},\hat{B},\hat{C},\hat{D})\|_{\ell_1,*}$. With this definition we obtain
		\begin{align}
		    \bE[|V_N(f)-\hat{\mathcal{L}}(f)|]\leq \frac{2}{N}\sqrt{\mu_{\max}(Q_e)}G_1(f)G_{-1}(f)G_2(f)\sqrt{K_w} \sum_{k=0}^{\infty} \|\hat{A}^k\|_2
		\end{align}
		Finally, with $G_0(f)= \sum_{k=0}^{\infty} \| \hat{A}^k \|_2$ and $G_3(\w)=\sqrt{\mu_{\max}(Q_e)K_w}$, we obtain the statement of the theorem
		\begin{align}
		    \bE[|V_N(f)-\hat{\mathcal{L}}(f)|]\leq \frac{2}{N}G_1(f)G_{-1}(f)G_2(f)G_3(\w) G_0(f)
		\end{align}
	}
\end{proof}
\begin{corollary}\label{cor:PV<hL} With notation as above the following holds
\begin{equation}
    \bP\left ( \Efrho V_N(f) \leq \Efrho \hat{\mathcal{L}}_N(f) + \frac{2}{\delta N}\Efrho G(f)\right ) > 1-\delta 
\end{equation}
\end{corollary}
\begin{proof}[Proof of Corollary \ref{cor:PV<hL}]
    We can apply Markov inequality to random variable $|V_N(f)-\hat{\mathcal{L}}_N(f)|$
    \begin{equation}
        \bP(|V_N(f)-\hat{\mathcal{L}}_N(f)| < \delta^{-1}\bE[|V_N(f)-\hat{\mathcal{L}}_N(f)|]) > 1-\delta
    \end{equation}
    We can use Lemma \ref{lem:Ev-hl}, and the fact that $V_N(f)-\hat{\mathcal{L}}_N(f)\leq |V_N(f)-\hat{\mathcal{L}}_N(f)|$, to obtain
    \begin{equation}
        \bP\left ( V_N(f)-\hat{\mathcal{L}}_N(f) < \frac{2}{\delta N}G(f) \right ) > 1-\delta
    \end{equation}
    Moving $\hat{\mathcal{L}}_N(f)$ to the right hand side of the inequality, and taking expectations over distributions $\rho(f)$, we obtain the statement of the lemma
    \begin{equation*}
        \bP\left ( \Efrho V_N(f) \leq \Efrho \hat{\mathcal{L}}_N(f) + \frac{2}{\delta N}\Efrho G(f)\right ) > 1-\delta 
    \end{equation*}
\end{proof}
\begin{lemma} Let $\sigma(r)$, be such that the following holds.
	\begin{align}
		\sigma(r)&\geq \sup_{t,k,l}\bE[\|\e(t,k,l)\|_2^r]\\
		\e(t,k,j)&=\begin{cases} Q_e-\e_g(t-k)\e_g^T(t-j),& k=j\\ -\e_g(t-k)\e_g^T(t-j),& k\neq j \end{cases}
	\end{align}
	Then the raw moments are bounded \label{lemma:L-Vr}
	\begin{align} 
		\bE[(\mathcal{L}(f)-&V_N(f))^r]\leq\frac{1}{N}\sigma(r)4(r-1)G_e(f)^{2r}
	\end{align}
\end{lemma}
\begin{proof}[Proof of Lemma \ref{lemma:L-Vr}]
	
	The prediction error can be expressed as
	\begin{align*}
		(\y(t)-\hyfi(t))=\sum_{k=0}^\infty\alpha_k\e_g(t-k)
	\end{align*}
	with
	\begin{align*}
		\alpha_k=\begin{cases} D_{e},& k=0\\ C_{e}A_e^{k-1}K_e,& k>0 \end{cases}
	\end{align*}
	Then generalised loss $\mathcal{L}(f)$ is expressed as
	\begin{align*}
		\mathcal{L}(f)&=\bE[(\y(t)-\hyfi(t))^2]\\
		&=\bE\left[\text{trace}\left ( \left(\sum_{k=0}^\infty\alpha_k\e_g(t-k)\right)\left(\sum_{k=0}^\infty\alpha_k\e_g(t-k)\right)^T \right )\right]\\
		&=\sum_{k=0}^\infty\alpha_kQ_e\alpha_k^T
	\end{align*}
	and infinite horizon prediction loss is 
	\begin{align*}
		V_N(f)&=\frac{1}{N}\sum_{k=0}^{N-1}(\y(t)-\hyfi(t))^2\\
		\mathcal{L}(f)-V_N(f)&=\frac{1}{N}\sum_{t=0}^{N-1}\left (\sum_{k=0}^\infty\alpha_kQ_e\alpha_k^T  - \sum_{k=0}^\infty\sum_{j=0}^{\infty}\alpha_k\e_g(t-k)\e_g(t-j)\alpha_k^T \right )\\
		&=\frac{1}{N}\sum_{k=0}^\infty\sum_{j=0}^{\infty}\alpha_k\e(t,k,j)\alpha_j^T \\
		\e(t,k,j)&=\begin{cases} Q_e-\e_g(t-k)\e_g^T(t-j),& k=j\\ -\e_g(t-k)\e_g^T(t-j),& k\neq j \end{cases}
	\end{align*}
	For ease of notation let us define
	\begin{align*}
		\z(t,k,j)=\alpha_k\e(t,k,j)\alpha_j
	\end{align*}
	then
	\begin{align*}
		&\bE[(\mathcal{L}(f)-V_N(f))^r]\\
		& = \frac{1}{N^r} \sum_{t_1=0}^{N-1}\dots \sum_{t_r=0}^{N-1} \sum_{k_1,j_1=0}^\infty \dots \sum_{k_r,j_r=0}^\infty \bE\left [\prod_{l=1}^{r}z(t_l,k_l,j_l)\right]
	\end{align*}
	Note that, with i.i.d. innovation noise $\e_g(t)$, if
	\begin{align*}
		&t_r-k_r\notin \{ t_i-k_i,t_i-j_i\}_{i=1}^{r-1}\\
		&\quad \land t_r-j_r\notin \{ t_i-k_i,t_i-j_i\}_{i=1}^{r-1}
	\end{align*}
	or similarly
	\begin{equation}
	\label{pf:b8:eq1}
		\{t_r-k_r,t_r-j_r\} \cap \{ t_i-k_i,t_i-j_i\}_{i=1}^{r-1} = \emptyset
	\end{equation}
	then $\z(t_r,k_r,j_r)$ is independent of $\z(t_i,k_i,j_i)$. Moreover,
	notice that $E(\z(t_r,k_r,j_r)]=0$. 
	Hence, if \eqref{pf:b8:eq1}, it holds that
	\begin{equation}
	\label{pf:b8:eq2}
		\bE\left [\prod_{l=1}^{r}z(t_l,k_l,j_l)\right] = \bE\left [\prod_{l=1}^{r-1}\z(t_l,k_l,j_l)\right ]\underset{=0}{\underbrace{\bE[\z(t_r,k_r,j_r)]}}=0.
	\end{equation}
	Let us denote
	\begin{align*}
		\mathcal{Z}=\{ t_i-k_i+k_r,t_i-j_i+k_r,t_i-k_i+j_r,t_i-j_i+j_r\}_{i=1}^{r-1}.
	\end{align*}
	Then using \eqref{pf:b8:eq2} for those $\{t_l,k_l,j_l\}_{l=1}^{r}$ which satisfy
	\eqref{pf:b8:eq1}, it follows that
	\begin{equation}
	\label{pf:b8:eq22}
		\bE[(\mathcal{L}(f)-V_N(f))^r]=\frac{1}{N^r}\sum_{t_1=0}^{N-1}\dots \sum_{t_{r-1}=0}^{N-1}\sum_{k_1,j_1=0}^\infty \dots \sum_{k_r,j_r=0}^\infty \sum_{t_{r}\in\mathcal{Z}}\bE\left [\prod_{l=1}^{r}z(t_l,k_l,j_l)\right ].
	\end{equation}
	Note that 
	\begin{align*}
		\bE\left [\prod_{l=1}^{r}z(t_l,k_l,j_l)\right ]&\leq \left |\bE\left [\prod_{l=1}^{r}z(t_l,k_l,j_l)\right ] \right |\leq \bE\left [\prod_{l=1}^{r}|z(t_l,k_l,j_l)|\right ].
	\end{align*}
	Let us focus on $|\z(t_i,k_i,j_i)|$:
	\begin{align*}
		|\z(t_l,k_l,j_l)| &\leq \|\alpha_{k_l}\|_2\|\alpha_{j_l}\|_2\|\e(t_l,k_l,j_l)\|_2\\
		\bE\left [\prod_{l=1}^{r}|\z(t_l,k_l,j_l)|\right ]&\leq \prod_{l=1}^r\|\alpha_{k_l}\|_2\|\alpha_{j_l}\|_2\bE\left [\prod_{l=1}^r\|\e(t_l,k_l,j_l)\|_2 \right ]
	\end{align*}
	Then using Arithmetic Mean-Geometric Mean Inequality, \cite{steele2004cauchy} we have
	\begin{align}
		\bE\left [\prod_{l=1}^r|\e(t_l,k_l,j_l)| \right ] \leq \frac{1}{r}\sum_{l=1}^r \bE[\|\e(t_l,k_l,j_l)\|_2^r] \label{pf:b8:eq3}
	\end{align}
	Now, let $\sigma(r)$, be such that the following holds.
	\begin{align}
		\sigma(r)\geq \sup_{t,k,l}\bE[\|\e(t,k,l)\|_2^r]
	\end{align}
	Then, 
	\(	\frac{1}{r}\sum_{l=1}^r \bE[\|\e(t_l,k_l,j_l)\|_2^r] \leq \sigma(r) \)
	 and then from \eqref{pf:b8:eq3} it follows that 
	\begin{align}
		\bE\left [\prod_{l=1}^r|\e(t_l,k_l,j_l)| \right ] \leq \sigma(r)
	\end{align}
	Combining this with \eqref{pf:b8:eq22}, it follows that
	\begin{align}
		&\bE[(\mathcal{L}(f)-V_N(f))^r]\leq\frac{1}{N^r}\sum_{t_1=0}^{N-1}\dots \sum_{t_{r-1}=0}^{N-1}\sum_{k_1,j_1=0}^\infty \dots \sum_{k_r,j_r=0}^\infty \sum_{t_{r}\in\mathcal{Z}}\sigma(r)\prod_{l=1}^r\|\alpha_{k_l}\|_2\|\alpha_{j_l}\|_2
		\label{pf:b8:eq4}
	\end{align}
	and the quantity $\sigma(r)\prod_{l=1}^r\|\alpha_{k_l}\|_2\|\alpha_{j_l}\|_2$ does not depend on $t_r$. Moreover 
	\begin{align*}
		\sum_{t_{r}\in\mathcal{Z}}\sigma(r)\prod_{l=1}^r\|\alpha_{k_l}\|_2\|\alpha_{j_l}\|_2\leq
		\sigma(r)\prod_{l=1}^r\|\alpha_{k_l}\|_2\|\alpha_{j_l}\|_2 |\mathcal{Z}|,
	\end{align*}
	where $|\mathcal{Z}|$ is the cardinality of the set $\mathcal{Z}$. Note $|\mathcal{Z}|\leq 4(r-1)$, therefore
	\begin{align*}
		\sum_{t_{r}\in\mathcal{Z}}&\sigma(r)\prod_{l=1}^r\|\alpha_{k_l}\|_2\|\alpha_{j_l}\|_2\leq \sigma(r)\prod_{l=1}^r\|\alpha_{k_l}\|_2\|\alpha_{j_l}\|_2 4(r-1),
	\end{align*}
	Combining the latter inequality with \eqref{pf:b8:eq4}, it follows that
	\begin{align}
		\bE[(\mathcal{L}(f)-V_N(f))^r]&\leq \frac{1}{N^r}\sum_{t_1=0}^{N-1}\dots \sum_{t_{r-1}=0}^{N-1}\sigma(r)4(r-1)\sum_{k_1,j_1=0}^\infty \dots \sum_{k_r,j_r=0}^\infty  \prod_{l=1}^r\|\alpha_{k_l}\|_2\|\alpha_{j_l}\|_2 
	\end{align}
	Now notice
	\begin{align*}
		G_e(f)^{2r}=\left ( \sum_{k=0}^\infty \|\alpha_k\|_2 \right )^{2r}=\left ( \sum_{k,j=0}^\infty \|\alpha_k\|_2\|\alpha_j\|_2 \right )^r \\
		=\sum_{k_1,j_1=0}^\infty \dots \sum_{k_r,j_r=0}^\infty  \prod_{l=1}^r\|\alpha_{k_l}\|_2\|\alpha_{j_l}\|_2 
	\end{align*}
	therefore we obtain
	\begin{align*}
		\bE[(\mathcal{L}(f)-V_N(f))^r] &\leq \frac{1}{N^r}\sum_{t_1=0}^{N-1}\dots \sum_{t_{r-1}=0}^{N-1}\sigma(r)4(r-1)G_e(f)^{2r} \\
		&\leq\frac{1}{N^r}N^{r-1}\sigma(r)4(r-1)G_e(f)^{2r}\\
		&\leq\frac{1}{N}\sigma(r)4(r-1)G_e(f)^{2r}\\
	\end{align*}
\end{proof}

\begin{lemma}\label{lemma:sigmar} 
For $r \ge 2$, the quantity $\sigma(r)$,
	\begin{align*}
		\sigma(r)=\max \left \{(\mu_{\max}(Q_e)^r4(m+r-1)!), (\mu_{\max}(Q_e)^{r}3^r(m+r-1)!)  \right \} = \mu_{\max}(Q_e)^{r}3^r(m+r-1)!
	\end{align*}
satisfies
	\begin{align*}
		\sigma(r)\geq \sup_{t,k,l}\bE[\|\e(t,k,l)\|_2^r]
	\end{align*}
\end{lemma}

\begin{proof}[Proof of Lemma \ref{lemma:sigmar}]
	Recall that
	\begin{align*}
		\e(t,k,j)=\begin{cases} Q_e-\e_g(t-k)\e_g^T(t-j),& k=j\\ -\e_g(t-k)\e_g^T(t-j),& k\neq j \end{cases}
	\end{align*}
	First let us take the case when $k\neq j$. Then
	\begin{align*}
		\bE[\|\e(t,k,l)\|_2^r]=\bE[\|-\e_g(t-k)\e_g^T(t-j)\|_2^r]
	\end{align*}
	Again as $\e_g(t)$ is i.i.d. we have 
	\begin{align*}
		\bE[\|\e(t,k,l)\|_2^r]\leq \bE[\|\e_g(t-k)\|_2^r]\bE[\|\e_g(t-j)\|_2^r]
	\end{align*}
	and due to stationarity of $\e_g(t)$, we have $\bE[\|\e_g(t-k)\|_2^r]=\bE[\|\e_g(t-j)\|_2^r]$, therefore
	\begin{align*}
		\bE[\|\e(t,k,l)\|_2^r]\leq \bE[\|\e_g(t)\|_2^r]^2
	\end{align*}
	and again due to stationarity of $\e_g(t)$, the moments do not depend on $t$, and using Lemma \ref{lemma:Emoments} we obtain
	\begin{align*}
		\sigma(r)\geq \mu_{\max}(Q_e)^r4((m+r-1)!) \geq \bE[\|\e(t,k,l)\|_2^r]^2
	\end{align*}
	Now let us take the case when $k=j$. Then
	\begin{align*}
		\bE[\|\e(t,k,l)\|_2^r]&=\bE[\|Q_e-\e_g(t-k)\e_g^T(t-k)\|_2^r]\\
		&\leq \bE[(\|Q_e\|_2+\|\e_g(t)\|_2^2)^r]\\
		&= \bE \left [ \sum_{j=0}^r \begin{pmatrix} r\\ j \end{pmatrix} \|Q_e\|_2^{r-j}\|\e_g(t)\|_2^{2j} \right ]\\
		&=\sum_{j=0}^r \begin{pmatrix} r\\ j \end{pmatrix} \|Q_e\|_2^{r-j}\bE\|\e_g(t)\|_2^{2j}]
	\end{align*}
	As $Q_e$ is a positive definite matrix,$\|Q_e\|_2 = \mu_{max}(Q_e)$, and hence
	\begin{align*}
		\bE[\|\e(t,k,l)\|_2^r]\leq \sum_{j=0}^r \begin{pmatrix} r\\ j \end{pmatrix} \mu_{\max}(Q_e)^{r-j}\bE\|\e_g(t)\|_2^{2j}]
	\end{align*}
	using Lemma \ref{lemma:evenEmoments} we obtain
	\begin{align*}
		\bE[\|\e(t,k,l)\|_2^r]&\leq \sum_{j=0}^r \begin{pmatrix} r\\ j \end{pmatrix} \mu_{\max}(Q_e)^{r-j}\mu_{\max}(Q_e)^{j}2^j(m+j-1)!\\
		&\leq \mu_{\max}(Q_e)^{r} \sum_{j=0}^r \begin{pmatrix} r\\ j \end{pmatrix}2^j(m+j-1)!.
	\end{align*}
	Since for $j\leq r$,  $(m+j-1)!\leq (m+r-1)!$, hence
	\begin{align*}
		&\bE\|\e(t,k,l)\|_2^{2r}]\leq \mu_{\max}(Q_e)^{r}(m+r-1)! \sum_{j=0}^r \begin{pmatrix} r\\ j \end{pmatrix}2^j
	\end{align*}
	Notice $3^r=(1+2)^r=\sum_{j=0}^r \begin{pmatrix} r\\ j \end{pmatrix}2^j$, hence
	\begin{align*}
		\bE\|\e_g(t,k,l)\|_2^{2r}]\leq \mu_{\max}(Q_e)^{r}3^r(m+r-1)!
	\end{align*}
	Hence,
	\begin{align*}
		\sigma(r)=\max \left \{\mu_{\max}(Q_e)^r4(m+r-1)!, \right . \\
		\left .\mu_{\max}(Q_e)^{r}3^r(m+r-1)!  \right \}.
	\end{align*}
	As we are interested in moments higher or equal to two, i.e. $r\geq 2$, then
	\begin{align*}
		\sigma(r)=\mu_{\max}(Q_e)^{r}3^r(m+r-1)!.
	\end{align*}
\end{proof}
\begin{lemma}\label{lem:mgf} For $\lambda\leq \left ( 3(m+1) \mu_{\max}(Q_e)G_e(f)^{2} \right )^{-1}$, the moment generating function is bounded
        \begin{align}
            		\bE\left [e^{\lambda(\mathcal{L}(f)-V_N(f))} \right ]\leq 1+\frac{2}{N}\frac{(m+1)! \left (3\lambda\mu_{\max}(Q_e)G_e(f)^{2}\right )^2}{(1-3(m+1)\lambda\mu_{\max}(Q_e)G_e(f)^{2})}
        \end{align}
\end{lemma}
\begin{proof}[Proof of Lemma \ref{lem:mgf}] \label{proof:mgf} 
	We can bound the moment generating function via series expansion. First note that $\bE[\mathcal{L}(f)-V_N(f)]=0$, and hence 
	\begin{align*}
		\bE\left [e^{\lambda(\mathcal{L}(f)-V_N(f))} \right ]=1+\lambda\bE[\mathcal{L}(f)-V_N(f)]+\sum_{r=2}^\infty \frac{\lambda^r}{r!}E[(\mathcal{L}(f)-V_N(f))^r].
	\end{align*}
	 Then using Lemma \ref{lemma:L-Vr} we get
	\begin{align}
		\bE\left [e^{\lambda(\mathcal{L}(f)-V_N(f))} \right ]\leq 1+\sum_{r=2}^\infty \frac{\lambda^r}{r!}\frac{1}{N}\sigma(r)4(r-1)G_e(f)^{2r} 
	\end{align}
	Now using Lemma \ref{lemma:sigmar} we obtain
	\begin{align*}
		&\bE\left [e^{\lambda(\mathcal{L}(f)-V_N(f))} \right ]\leq 1+\frac{1}{N}\sum_{r=2}^\infty \frac{(m+r-1)!}{r!}4(r-1)\left (3\lambda\mu_{\max}(Q_e)G_e(f)^{2}\right )^r\\
	\end{align*}
	Notice that $4(r-1)\leq 2^r$, for $r\in\mathbb{N}$. Furthermore
	\begin{align*}
		\frac{(m+r-1)!}{r!}=m!\frac{m+1}{2}\frac{m+2}{3}\dots \frac{m+r-1}{r}
	\end{align*}
	and as $\frac{m+r-1}{r}\leq \frac{m+1}{2}$, for all $r\geq 2$, then
	\begin{align*}
		\frac{(m+r-1)!}{r!}\leq m!\left (\frac{m+1}{2} \right )^{r-1} = m!\frac{\left (\frac{m+1}{2} \right )^{r}}{\frac{m+1}{2}}		= 2\frac{m!}{m+1}\left (\frac{m+1}{2} \right )^r.
	\end{align*}
	 Hence, we cand derive the following inequality:
	\begin{align*}
		&\bE\left [e^{\lambda(\mathcal{L}(f)-V_N(f))} \right ]\leq 1+\frac{2}{N}\frac{m!}{m+1} \sum_{r=2}^\infty \left (3(m+1)\lambda\mu_{\max}(Q_e)G_e(f)^{2}\right )^r.
	\end{align*}
	Notice that if $$| (m+1)\lambda\mu_{\max}(Q_e)3\|f_e\|^{2} | < 1,$$ then the 
	infinite sum 
	$\sum_{r=2}^\infty \left (3(m+1)\lambda\mu_{\max}(Q_e)G_e(f)^{2}\right )^r$
	is absolutely convergent, and 
	\[
	  \begin{split}
	   & \sum_{r=2}^\infty \left (3(m+1)\lambda\mu_{\max}(Q_e)G_e(f)^{2}\right )^r 
	   = \frac{ \left (3(m+1)\lambda\mu_{\max}(Q_e)G_e(f)^{2}\right )^2}{1-3(m+1)\lambda\mu_{\max}(Q_e)G_e(f)^{2}}
	  \end{split}  
	\]
	 To sum up, if
	\begin{align*}
		\lambda\leq \left ( 3(m+1) \mu_{\max}(Q_e)G_e(f)^{2} \right )^{-1}.
	\end{align*}
	then 
	\begin{align*}
		\bE\left [e^{\lambda(\mathcal{L}(f)-V_N(f))} \right ]&\leq 1+\frac{2}{N}\frac{m!}{m+1} \frac{ \left (3(m+1)\lambda\mu_{\max}(Q_e)G_e(f)^{2}\right )^2}{1-3(m+1)\lambda\mu_{\max}(Q_e)G_e(f)^{2}}\\
		&\leq 1+\frac{2}{N}\frac{(m+1)! \left (3\lambda\mu_{\max}(Q_e)G_e(f)^{2}\right )^2}{(1-3(m+1)\lambda\mu_{\max}(Q_e)G_e(f)^{2})}.
	\end{align*}
	
\end{proof}
\begin{proof}[Proof of Theorem \ref{thm:pac:altKL}] \label{proof:pac:alt1} 
	Let us apply the Donsker \& Varadhan variational 
	formula to the function $\lambda(\mathcal{L}(f)-V_N(f))$ and applying Corollary \ref{cor:PV<hL}
	it then follows that
	with probability at least $1-2\delta$, the following 
	holds
	\begin{align}
		&E_{f\sim \hat{\rho}} \mathcal{L} (f) \le \  E_{f\sim \hat{\rho}} \hat{\mathcal{L}}_{N}(f) +\frac{2}{\delta N}E_{f\sim \hat{\rho}} G(f) +\dfrac{1}{\lambda}\!\left[KL(\hat{\rho} \|\pi) + \ln\dfrac{1}{\delta}	+ \Psi_{\pi}(\lambda,n) \right ],\label{T:pacAlt:eq1}
	\end{align}
    with
    \begin{equation}
        \Psi_{\pi}(\lambda,n)=	\ln E_{f\sim\pi} \bE[e^{\lambda(\mathcal{L}(f)-V_N(f))}]
    \end{equation}
    Using  Lemma \ref{lem:mgf}, we obtain
    \begin{align}
        \Psi_{\pi}(\lambda,n) &\le \ln \left(E_{f\sim\pi} \left[ 1+\frac{2}{N}\frac{(m+1)!  
	     \left (3\lambda\mu_{\max}(Q_e)G_e(f)^2 \right )^2}{(1-3(m+1)\lambda\mu_{\max}(Q_e)G_e(f)^{2})}\right] \right)\\
	     &=\ln \left(1+\frac{2}{N}E_{f\sim\pi} \left[\frac{(m+1)!  
	     \left (3\lambda\mu_{\max}(Q_e)G_e(f)^2 \right )^2}{(1-3(m+1)\lambda\mu_{\max}(Q_e)G_e(f)^{2})}\right] \right)
    \end{align}
    Let us define $G_e(\Theta)=\sup_{\theta\in\Theta}G_e(\theta)$, then
    \begin{align}
         \Psi_{\pi}(\lambda,n) &\le \ln \left(1+\frac{2}{N} \frac{(m+1)!  
	     \left (3\lambda\mu_{\max}(Q_e)G_e(\Theta)^2 \right )^2}{(1-3(m+1)\lambda\mu_{\max}(Q_e)G_e(\Theta)^{2})} \right) \label{eq:totalBoundonPsi}
    \end{align}
    Using \eqref{eq:totalBoundonPsi} in \eqref{T:pacAlt:eq1}, yields the results of theorem \ref{thm:pac:altKL}.
\end{proof}
\begin{proof}[Proof of Theorem \ref{thm:pac:alt1}] \label{proof:ofThm42}
    We start with Renyi change of measure \cite{pacRenyi}: 
    for any measurable function $\phi$,
	\begin{align}
		\frac{\alpha}{\alpha-1}\ln\Efrho \phi(f)\leq \mathcal{D}_\alpha(\rho \mid \mid \pi)+\ln\left (\Efpi \phi^\frac{\alpha}{\alpha-1}(f) \right ),
	\end{align}
	where $\mathcal{D}_\alpha(\rho \mid \mid \pi)$ is the Renyi divergence
	\[\mathcal{D}_\alpha(\rho \mid \mid \pi)=\frac{1}{\alpha-1}\ln\left ( \Efpi \left ( \frac{\rho(f)}{\pi(f)} \right )^\alpha \right ).\]
	With some rearrangement of the terms,  we obtain
	\begin{align}
		\Efrho \phi(f) \leq \left ( \Efpi \left ( \frac{\rho(f)}{\pi(f)} \right )^\alpha \right )^\frac{1}{\alpha}\left ( \Efpi \phi^\frac{\alpha}{\alpha-1}(f) \right )^\frac{\alpha-1}{\alpha}.
	\end{align}
	By choosing $\phi(f) \triangleq  \mathcal{L}(f)-V_N(f))$ and choosing $\alpha$ so that $r=\frac{\alpha}{\alpha-1}$ is even, i.e. $\alpha=\frac{r}{r-1}$, we obtain
	\begin{align}
		\Efrho \mathcal{L}(f)-v_N(f)) \leq \left ( \Efpi \left ( \frac{\rho(f)}{\pi(f)} \right )^\frac{r}{r-1} \right )^\frac{r-1}{r}\left ( \Efpi (\mathcal{L}(f)-v_N(f))^{r} \right )^\frac{1}{r}\\
		\label{eq:boundPreMarkov}
	\end{align}
	By applying Markov's inequality to $(\mathcal{L}(f)-\hat{\mathcal{L}}_N(f))^{r}$ 
	\begin{align}
		\bP((\mathcal{L}(f)-v_N(f))^{r} < \delta^{-1}\bE[(\mathcal{L}(f)-V_N(f))^{r}])>1-\delta \label{eq:markov}
	\end{align}
	applying \eqref{eq:markov} to \eqref{eq:boundPreMarkov}, the following holds with probability $1-\delta$
		\begin{align}
		 \Efrho (\mathcal{L}(f)-V_N(f)) \leq  \left ( \Efpi \left ( \frac{\rho(f)}{\pi(f)} \right )^\frac{r}{r-1} \right )^\frac{r-1}{r}\left ( \Efpi \delta^{-1}\bE[(\mathcal{L}(f)-V_N(f))^{r}] \right )^\frac{1}{r}\\
		 \Efrho (\mathcal{L}(f)-V_N(f)) \leq  \delta^{-\frac{1}{r}}\left ( \Efpi \left ( \frac{\rho(f)}{\pi(f)} \right )^\frac{r}{r-1} \right )^\frac{r-1}{r}\left ( \Efpi \bE[(\mathcal{L}(f)-V_N(f))^{r}] \right )^\frac{1}{r}.
	\end{align}
	Then by moving $\Efrho V_N(f)$ to the right hand side we obtain
	\begin{align}
		 \bP(\Efrho \mathcal{L}(f) &\leq \Efrho V_N(f) \nonumber \\
		  &\left . +  \delta^{-\frac{1}{r}}\left ( \Efpi \left ( \frac{\rho(f)}{\pi(f)} \right )^\frac{r}{r-1} \right )^\frac{r-1}{r}\left ( \Efpi \bE[(\mathcal{L}(f)-V_N(f))^{r}] \right )^\frac{1}{r}\right ) > 1-\delta 
		  \label{eq:boundPreMoment}
	\end{align}
	From Lemma \ref{lemma:L-Vr} and lemma \ref{lemma:sigmar}, we know that for $r\geq2$
	\begin{equation}
		\bE[(\mathcal{L}(f)-V_N(f))^{r}]\leq \frac{4}{N}\mu_{\max}(Q_e)^{r}3^{r}(m+r-1)!(2r-1)G_e^{2r}(f) \label{eq:L-Vrrecap}
	\end{equation}
	Taking \eqref{eq:L-Vrrecap} into \eqref{eq:boundPreMoment} we obtain
	\begin{align}
		& \bP \left( \Efrho \mathcal{L}(f) \leq \Efrho V_N(f) +  3\left (\frac{4}{\delta N}\right )^\frac{1}{r}\mu_{\max}(Q_e)((m+r-1)!(r-1))^\frac{1}{r} \times  \right.
		\nonumber \\
		&\left.  \cdot \left ( \Efpi \left ( \frac{\rho(f)}{\pi(f)} \right )^\frac{r}{r-1} \right )^\frac{r-1}{r}\left ( \Efpi G_e^{2r}(f) \right )^\frac{1}{r} \right) > 1-\delta \label{eq:boundPreLhat}
	\end{align}
	By combining \eqref{eq:boundPreLhat} and Corollary \ref{cor:PV<hL}, we obtain the statement of the theorem.
\end{proof}

\subsection{Multiple Output}
The results presented in the paper assume single output, i.e. $\y(t)\in\reals^1$. In order to generalise the results to multiple outputs, i.e. $n_y>1$, we introduce the following
notation. 
\\
Consider a predictor $f$ from $\mathcal{F}$. Denote by
$f_p:\mathcal{W}^{*} \rightarrow \mathbb{R}$ 
the predictor, such that $f_p(\underline{w})$
is the $p$th component of $f(\underline{w}) \in \mathbb{R}^{n_y}$
for all $\underline{w} \in \mathcal{W}^{*}$. Define
$\mathcal{F}_{p}=\{ f_p \mid f \in \mathcal{F}\}$
for all $p=1,\ldots,n_y$. 
It is easy to see that  
$\mathcal{F}_p$ also satisfies Assumption \ref{as:parameterisation} from
Section \ref{sect:learn}
\\
Moreover, let $\y_p(t)$ be the $p$th component of 
$\y(t)$.  We will argue that for any $f \in \mathcal{F}$,
the predictor
$f_p$ can be used to predict $\y_p$ based in $\w$ and
$\y_p$, $\w$ satisfies Assumption \ref{as:generator}. 
\\
If $\w=\mathbf{u}(t)$, then 
$\y_p$ and $\w$ satisfy Assumption \ref{as:generator} with $\y$ being replaced
by $\y_p$.  Moreover, in this case 
$f_p$ is a predictor for $\mathcal{W}=\mathbb{R}^{n_m}$.
\\
If $\w=\begin{bmatrix} \y^T & \mathbf{u}^T \end{bmatrix}^T$, then  Assumption \ref{as:generator} is satisfied
with $\y$ being replaced by $\y_p$ and 
$\mathbf{u}$ being replaced by 
$\mathbf{u}_{+,p}=\begin{bmatrix} \y_{-,p}^T, \mathbf{u}^T \end{bmatrix}^T$,
where $\y_{-,p}(t)$ is the vector obtained from $\y(t)$ by
leaving out its $p$th component.
In this case, $f_p$ can be interpreted as a predictor
acting on $\w$, by letting.
$\mathbf{u}_{+,p}$ play the role of $\mathbf{u}$.
Indeed, $\begin{bmatrix} \y_p^T & \mathbf{u}_{+,p}^T \end{bmatrix}^T$ can naturally be identified with
$\w=\begin{bmatrix} \y^T & \mathbf{u}^T \end{bmatrix}^T$,
after rearranging the order of the elements. 
\\
Moreover, notice that any density $\rho,\pi$ on $\mathcal{F}$
can naturally be interpreted as a density on $\mathcal{F}_p$.
\\
This means that Theorem \ref{thm:pac:altKL} and Theorem \ref{thm:pac:alt1} hold for
the hypothesis class $\mathcal{F}_p$ and any
density $\rho,\pi$ on $\mathcal{F}$, if the latter are 
interpreted as densities on $\mathcal{F}_p$.
\\
For any $f \in \mathcal{F}$, 
denote by $\mathcal{L}_p(f)$, $\hat{\mathcal{L}}_{N,p}(f)$
the generalization and empirical losses respectively for
the predictor $f_p$ and output process $\y_p(t)$. 
More precisely, let us denote the finite and infinite past predictions 
generated by $f_p$ by
$\hat{\y}_{f,p}(t \mid 0)$ and  $\hat{\y}_{f,p}(t)$
respectively. In particular, 
$\hat{\y}_{f,p}(t \mid 0)$  and $\hat{\y}_{f,p}(t)$
are the $p$th component of the predictions 
$\hat{\y}_f(t \mid 0)$ and $\hat{\y}_f(t)$ respectively,
which are generated by $f$.
It then follows that  $\hat{\mathcal{L}}_{N,p}(f)=\frac{1}{N}\sum_{i=0}^{N-1} (\hat{\y}_{f,p}(i \mid 0)- \y_p(i))^2$, 
$\mathcal{L}_p(f)=\bE[(\hat{\y}_{f,p}(t)- \y_p(t))^2]$.
\\
Using the notation and discussion above, it follows that
for any two densities 
$\rho$ and $\pi$ on $\mathcal{F}$, and 
all $p=1,\dots, n_{\mathrm y}$, and for all $\delta_p > 0$,
\begin{align}
\label{thm:decompPACBound:eq1}
    \bP\Big (\omega\in\Omega\mid \Efrho \mathcal{L}_p(f) \leq \Efrho \hat{\mathcal{L}}_{N,p}(f)(\omega) + r_{N,p} \Big )\leq 1-\delta_p
\end{align}
holds, where
the error term $r_{N,p}$ is from Theorem \ref{thm:pac:altKL} or Theorem \ref{thm:pac:alt1} applied to $\mathcal{F}_{p}$, the
output $\y_p$ and the input $\mathbf{u}$ or $\mathbf{u}_{+,p}$ and $2\delta=\delta_p$.
\begin{Theorem}[Multiple outputs]\label{thm:decompPACbound}
If for any $p=1,\ldots,n_y$ \eqref{thm:decompPACBound:eq1} holds, then the following error bound holds:
\begin{align}
    \bP\Big (\omega\in\Omega\mid \Efrho \mathcal{L}(f) \leq \Efrho \hat{\mathcal{L}}_N(f)(\omega) + \sum_{p=1}^{n_y}r_{N,p} \Big ) \leq 1-\sum_{p=1}^{n_y}\delta_p
\end{align}
\end{Theorem}
For the PAC-Bayesian-Like bounds (Theorems \ref{thm:pac:altKL} and \ref{thm:pac:alt1}) proposed in the paper, we always have $\delta_p=2\delta$ and thus
\begin{align}
    \bP\Big (\omega\in\Omega\mid \Efrho \mathcal{L}(f) \leq \Efrho \hat{\mathcal{L}}_N(f)(\omega) + \sum_{p=1}^{n_y}r_{N,p} \Big ) \leq 1- 2n_y\delta \label{eq:pac:MO}
\end{align}
the error terms $r_{N,p}$ only differ in $D_{e,p}$ and $C_{e,p}$, where $D_{e,p}$ and $C_{e,p}$ are the $p$'th rows of matrices $C_e$ and $D_e$, defined in Definition \ref{def:constants}, now with $C_1,\hat{C},\hat{D}$ appropriate for $\y(t)\in\reals^{n_y}$, and $\hat{\y}_f(t)\in\reals^{n_y}$

\begin{proof}[Proof of Theorem \ref{thm:decompPACbound}]
first notice that the square loss can be decomposed as
\begin{align*}
    \mathcal{L}(f)&=\lim_{s\to -\infty}\bE[(\y(t)-\hyf(t|s))^T(\y(t)-\hyf(t|s))]\\
    &=\lim_{s\to -\infty}\bE\Big [\sum_{p=1}^{n_y} (\y_p(t)-\hat{\y}_{f,p}(t|s))^2 \Big ]\\
    &=\sum_{p=1}^{n_y} \lim_{s\to -\infty}\bE\Big [(\y_p(t)-\hat{\y}_{f,p}(t|s))^2 \Big ]=\sum_{p=1}^{n_y} \mathcal{L}_p(f)\\
    \Efrho \mathcal{L}(f) &=   \sum_{p=1}^{n_y} \Efrho \mathcal{L}_p(f) \\
    \hat{\mathcal{L}}_N(f) & = \frac{1}{N} \sum_{i=0}^{N} (\y(t)-\hyf(i \mid 0))^T(\y(t)-\hyf(i \mid 0)) = \nonumber 
    \\ 
    & = \frac{1}{N} \sum_{i=0}^{N} \sum_{p=1}^{n_y} (\y_p(t)-\hat{\y}_{f,p}(i \mid 0))^2=\sum_{p=1}^{n_y} \hat{\mathcal{L}}_{N,p}(f) \\
     \Efrho \hat{\mathcal{L}}_{N}(f) &=\sum_{p=1}^{n_y} \Efrho \hat{\mathcal{L}}_{N,p}(f)
\end{align*}
Now if we use the PAC-Bayesian-Like upper bound on each of the single output generalization loss $\mathcal{L}_p(f)$ we obtain
\begin{align}
    \sum_{p=1}^{n_y}\Efrho \mathcal{L}_p(f) \leq \sum_{p=1}^{n_y}\Efrho \hat{\mathcal{L}}_{N,p}(f) + r_{N,p}\\
    \Efrho \mathcal{L}(f) \leq \Efrho \hat{\mathcal{L}}_N(f) + \sum_{p=1}^{n_y}r_{N,p}
\end{align}
However, each $\Efrho \mathcal{L}_p(f)$ can only be upper bounded with probability less than $1-\delta_p$, and we need all upper-bounds to hold. Therefore we take intersections between the sets $S_p$
\begin{align}
    S_p=\{\omega\in\Omega\mid \Efrho \mathcal{L}_p(f) \leq  \Efrho \hat{\mathcal{L}}_{N,p}(f)(\omega) + r_{N,p}\}\\
    \bP(S_p)\leq 1-\delta_p,\\
    \bP(\bar{S_p})=\bP(\Omega\setminus S_p)\geq \delta_p
\end{align}
so that
\begin{align}
    \bP\Big (\omega\in\Omega\mid \Efrho \mathcal{L}(f) \leq \Efrho \hat{\mathcal{L}}_N(f)(\omega) + \sum_{p=1}^{n_y}r_{N,p} \Big )=\bP \Big (\bigcap_{p=1}^{n_y} S_p \Big )\leq 1- \bP \Big ( \bigcup_{p=1}^{n_y} \bar{S}_p \Big )
\end{align}
Since $\bP \Big ( \bigcup_{p=1}^{n_y} \bar{S}_p \Big )\leq \sum_{p=1}^{n_y} \bP \Big ( \bar{S}_p \Big )\leq\sum_{p=1}^{n_y}\delta_p$, we get the results of the theorem.\\

\end{proof}

\section{Predictors as stochastic LTI systems}
  As it was discussed in Section \ref{sect:LTIaspred} stochastic LTI systems give rise
  to optimal predictors and the problem of learning LTI systems boils down to finding
  optimal predictors of $\y(t)$ based on past and present values of $\y$ and $\mathbf{u}$. 
 \\
  Assume first that $\w(t)=\mathbf{u}(t)$.  As it was noted in Section \ref{sect:LTIaspred}, the stochastic
  LTI eq. \eqref{eq:assumedSys:innov} gives rise to the predictor $f_{(A,B,C,D)}$, and the infinite past prediction 
  $\hat{\y}_{f_{(A,B,C,D)}}(t)$ can be expressed as 
   \begin{equation}
  \label{pred2stoch:eq0}
   \begin{split}
    & \x_d(t+1)=A\x_d(t)+B\mathbf{u}(t) \\
    & \hat{\y}_{f_{(A,B,C,D)}}(t)=\y^d(t)=C\x_d(t)+D\mathbf{u}(t)
 \end{split}    
  \end{equation}
  Note that $\y(t)=\y^d(t)+\y^s(t)$, where 
  $\y^s(t)=\e(t)+\sum_{k=0}^{\infty} CA^{k}K \e(t-k)$, i.e.,
  the output 
  $\y^d(t)= \hat{\y}_{f_{(A,B,C,D)}}(t)$ of the predictor
  $f_{(A,B,C,D)}$ represents the part of the output which depends on
  the input $\mathbf{u}$, and $\y^s(t)$ represents the part of the output which depends on the noise.  If there is no feedback from $\y$ to $\mathbf{u}$ ( see \cite[Chapter 17]{LindquistBook} for the  definition of absence of feedback from $\y$ to $\mathbf{u}$), then 
  $\y^s(t)$ is in fact uncorrelated with the inputs $\{\mathbf{u}(s)\}_{s \in \mathbb{Z}}$ and
  $\y^d(t)=\bE[\y(t) \mid \{\mathbf{u}(s)\}_{s \in \mathbb{Z}}]$. 
  In the case of no feedback from $\y$  to $\mathbf{u}$, 
  the  generalization error of $f_{(A,B,C,D)}$ equals
  $\bE[\|\y^s(t)\|_2^2]=\bE[\|\y(t)-\bE[\y(t) \mid \{\mathbf{u}(s)\}_{s \in \mathbb{Z}}]]$, i.e., it is the best (smallest variance) linear estimate
  of $\y(t)$ using $\{\mathbf{u}(s)\}_{s \le t}$.
  \\
  Conversely, let $f=f_{(\hat{A},\hat{B},\hat{C},\hat{D})}$ be a predictor from
  the set $\mathcal{F}$.
  It then follows that that the infinite past prediction $\hat{\y}_f(t)$ can be expressed as
  \begin{equation}
  \label{pred2stoch:eq1}
    \hat{\x}(t+1)=\hat{A}\hat{\x}(t)+\hat{B}\w(t), ~~~ \hat{\y}_f(t)=\hat{C}\hat{\x}(t)+\hat{D}\w(t)
  \end{equation}
  where $\hat{\x}$ is the unique stationary process 
  which satisfies \eqref{pred2stoch:eq1}. 
  \\
  Assume that the stochastic LTI system eq. \eqref{eq:assumedSys:innov} from Section \ref{sect:LTIaspred} is such that
  $(A,B)$ is controllable, $(C,A)$ is observable, i.e., the deterministic LTI system $(A,B,C,D)$
  is minimal. Moreover assume that the LTI system $(\hat{A},\hat{B},\hat{C},\hat{D})$
  representing the predictor $f$ is also minimal, i.e., $(\hat{A},\hat{B})$ is controllable and $(\hat{C},\hat{A})$ is observable.
 \\
  If $\y^d(t)=\hat{\y}_f(t)$, then from \cite[Theorem 4.1]{picci1996stochastic} it follows that
  $D=\hat{D}$ and the deterministic LTI system
  $(A,B,C,D)$ and $(\hat{A},\hat{B},\hat{C})$ are  similar: there exists a nonsingular
  matrix $\mathcal{T}$ such that $A=\mathcal{T}\hat{A}\mathcal{T}^{-1}$,
  $B=\mathcal{T}\hat{B}$, $C=\hat{C}\mathcal{T}^{-1}$. 
  In particular, in this case $f$ is the predictor associated with the stochastic LTI system
  described in eq. \eqref{eq:assumedSys:innov} of Section \ref{sect:LTIaspred}, i.e., $f=f_{(A,B,C,D)}$.
  \\
  Assume that $f$ is a predictor represented by a deterministic LTI system 
  $(\hat{A},\hat{B},\hat{C},\hat{D})$ is such that its generalization loss is small.
  As before, we assume that $(\hat{A},\hat{B},\hat{C},\hat{D})$ and
  $(A,B,C,D)$ are both minimal. 
  In this case it can be shown that the $H_2$ distance between 
  the deterministic LTI systems $(A,B,C,D)$  and that of the predictor $(\hat{A},\hat{B},\hat{C},\hat{D})$ is small.
  \\
  Indeed, from the well-known formula for spectral densities of outputs of LTI systems driven by stochastic inputs \cite{LindquistBook}, it follows that
  \begin{equation}
  \label{eq:pred:var1}
  \bE[\|\y^d(t)-\hat{\y}_f(t)\|^2_2]=\frac{1}{2\pi} \mathrm{trace} \int_{-\pi}^{\pi} (H_1-H_2)(e^{i\omega})\Phi_{\w}(e^{i\omega}(H_1^{T}-H_2^{T})(e^{-i\omega})d\omega
 \end{equation} 
  where $i=\sqrt{-1}$ and $H_1(z)=C(zI-A)^{-1}B+D$ and $H_2(z)=\hat{C}(zI-\hat{A})^{-1}\hat{B}+\hat{D}$
  and $\Phi_w(z)$ is the spectral density of $\w$.
  Assume $\Phi_w(e^{i\omega}) > \mu_wI$, $\omega \in [-\pi,\pi]$ for some $\mu_w > 0$. 
  This is the case when $w$ is coercive, for instance, it satisfies Assumption \ref{as:generator}.
  It then follows using Parseval's equality and the well-known properties of
  $H_2$ norm, that 
  \begin{equation}
\label{eq:pred:var2}
\begin{split}
  & \frac{1}{2\pi} \mathrm{trace} \int_{-\pi}^{\pi} (H_1-H_2)(e^{i\omega})\Phi_{\w}(e^{i\omega})(H_1^{T}-H_2^{T})(e^{-i\omega})d\omega
  \ge  \\
  & \mu_{\w} \frac{1}{2\pi}  \mathrm{trace} \int_{-\pi}^{\pi}  (H_1-H_2)(e^{i\omega})(H_1^{T}-H_2^{T})(e^{-i\omega})d\omega=\\
  & \left (\|D-\hat{D}\|_F^2+\sum_{k=0}^{\infty} \|CA^{k}B-\hat{C}\hat{A}^{k}\hat{B}\|_F^2 \right) \mu_{w},
\end{split}
\end{equation}
where $\|\cdot\|_F$ denotes the Frobenius norm,
and therefore
\begin{equation}
\label{eq:pred:var3}
\bE[\|\y^d(t)-\hat{\y}_f(t)\|^2_2] \ge \left(\|D-\hat{D}\|_F^2+\sum_{k=0}^{\infty} \|CA^{k}B-\hat{C}\hat{A}^{k}\hat{B}\|_F^2 \right) \mu_{w}
\end{equation}
  Hence, if $E[\|\y^d(t)-\hat{\y}_f(t)\|^2]$ is small, then the square of the $H_2$ distance
  $\|D-\hat{D}\|_F^2+\sum_{k=0}^{\infty} \|CA^{k}B-\hat{C}\hat{A}^{k}\hat{B}\|_F^2$
  between the deterministic LTI systems $(A,B,C,D)$ and $(\hat{A},\hat{B},\hat{C},\hat{D})$ is also small.
  This means that the output response of the stochastic LTI system \eqref{eq:assumedSys:innov} from Section \ref{sect:LTIaspred} is close to \eqref{pred2stoch:eq1} for any input $\mathbf{u}$.
  In particular, from \cite[Theorem 3.2.1]{RalfPeeters}, using minimality of
  $(A,B,C,D)$ and $(\hat{A},\hat{B},\hat{C},\hat{D})$, it follows that 
  if $\|D-\hat{D}\|_F^2+\sum_{k=0} \|CA^{k}B-\hat{C}\hat{A}^{k}\hat{B}\|_F^2$ is sufficiently small,
  then for a suitable  nonsingular matrix $\mathcal{T}$, the norms of the differences
  $\|\hat{A}-\mathcal{T}^{-1}A\mathcal{T}\|_F$,
  $\|\hat{B}-\mathcal{T}^{-1}B\|_F$, $\|\hat{C}-C\mathcal{T}\|_F$, $\|D-\hat{D}\|_F$ are small.
  Indeed, if $\|D-\hat{D}\|_F^2+\sum_{k=0} \|CA^{k}B-\hat{C}\hat{A}^{k}\hat{B}\|_F^2$ is sufficiently small, then
  by \cite[Theorem 3.2.1]{RalfPeeters}, the equivalence classes 
  of deterministic LTI systems isomorphic to $(A,B,C,D)$ and to
  $(\hat{A},\hat{B},\hat{C},\hat{D})$ respectively are close in
  the topology the manifold of minimal stable systems. 
  In particular, they belong to the same coordinate chart of this manifold,
  which means that the matrices of the isomorphic copies of $(A,B,C,D)$ and $(\hat{A},\hat{B},\hat{C},\hat{D})$ are close. 
  Since  replacing $A,B,C,K$ by $\mathcal{T}^{-1}A\mathcal{T}$,
  $\mathcal{T}^{-1}B$, $C\mathcal{T}$, $\mathcal{T}^{-1}K$ in \eqref{eq:assumedSys:innov} of Section \ref{sect:LTIaspred} 
  also results in a stochastic LTI system representation of $\y$, we can 
  view, without loss of generality,  the matrices $\hat{A},\hat{B},\hat{C},\hat{D}$
  as approximations of the matrices $A,B,C,D$ of eq. \eqref{eq:assumedSys:innov}.
  \\
  Assume now  that  $\w=\begin{bmatrix} \y^T & \mathbf{u}^T \end{bmatrix}^T$. As it was noted in Section \ref{sect:LTIaspred}, the stochastic system eq. \eqref{eq:assumedSys:innov} can be be associated with the predictor: 
  \begin{equation}\label{eq:assumedSys:pred}
 \begin{split}
	&	\x(t+1)=\underbrace{(A-KC)}_{\hat{A}_0}\x(t)+\underbrace{(B-KD)\mathbf{u}(t)+K\y(t)}_{\hat{B}_0\w(t)},\\
	&	\hat{\y}(t)=\underbrace{C}_{\hat{C}_0} \x(t)+\underbrace{D\mathbf{u}(t)}_{\hat{D}_0\w(t)}
\end{split}
\end{equation}
where $\hat{A}_0=A-KC$, $\hat{B}_0=\begin{bmatrix} K & B-KD \end{bmatrix}$,
$\hat{C}_0=C$, $\hat{D}_0=\begin{bmatrix} 0 & D \end{bmatrix}$. 
In addition, if there is no feedback from $\y$ to $\mathbf{u}$  (see \cite[Chapter 17]{LindquistBook} for definition), it can be shown that
the the predictor $f_0=f_{(\hat{A}_0,\hat{B}_0,\hat{C}_0,\hat{D}_0)}$
has the smallest generalization loss, i.e., 
$E[\|\y(t)-\hat{\y}_{f_0}(t)\|^2]$ is the smallest one among all the infinite
past prediction errors $E[\|\y(t)-\hat{\y}_{f}(t)\|^2]$, $f \in \mathcal{F}$.
\\
In particular, in the case of no feedback from $\y$ to $\mathbf{u}$, if $\hat{f}=\mathrm{arg min}_{f \in \mathcal{F}} E[\|\y(t)-\hat{\y}_{f}(t)\|^2]$,
and $\hat{f}=f_{(\hat{A},\hat{B},\hat{C},\hat{D})}$, then $\e(t)=\y(t)-\hat{\y}_{\hat{f}}(t)$ is the 
innovation process of $\y$ w.r. to the past outputs and past and present inputs, as defined in \cite[eq. (17.16))]{LindquistBook}, i.e., $\hat{\y}_{\hat{f}}(t)=E[\y(t) \mid \{\y(s),\mathbf{u}(s)\}_{s < t} \cup \{ \mathbf{u}(t)\}]$. 
It then follows that
\begin{equation}
\label{eq:pred2sys}
\begin{split}
& \x(t+1)=(\hat{A}+\hat{K}\hat{C})\x(t)+(\hat{B}_{u}+\hat{K}\hat{D}_u)\mathbf{u}(t)+\hat{K}\e(t) \\
& \y(t)= \hat{C}\x(t)+\hat{D}_{u}\mathbf{u}(t)+\e(t)
\end{split}
\end{equation}
where $\hat{B}=\begin{bmatrix} \hat{K} & \hat{B}_u \end{bmatrix}$ and
$\hat{D}=\begin{bmatrix} 0 & \hat{D}_u \end{bmatrix}$, is a stochastic LTI representation of
$\y$.
\\
Assume that the matrices of $(A,B,C,K,D)$ of the stochastic system from eq. \eqref{eq:assumedSys:innov},
Section \ref{sect:LTIaspred} satisfy the following conditions: the pair $(C,A)$ is observable, the pairs
$(A, K)$ and $(A,B)$ are controllable and assume that the covariance matrix of $\e(t)$ is strictly positive definite. Moreover, assume that the deterministic LTI system 
$(\hat{A},\hat{B},\hat{C},\hat{D})$ is such that the pairs 
$(\hat{A},\hat{B}_u)$ and $(\hat{A},\hat{K})$ are controllable and the pair
$(\hat{C},\hat{A})$  is observable. Furthermore, let us assume that there is no feedback from $\y$ to $\mathbb{u}$
Then from \cite{picci1996stochastic} it follows that
for a suitable nonsingular matrix  $\mathcal{T}$, $\hat{A}=\mathcal{T}A\mathcal{T}^{-1}$,
$\hat{B}_u=\mathcal{T}B$, $\hat{K}=\mathcal{T}K$, $\hat{C}=C\mathcal{T}^{-1}$, $\hat{D}_u=D$.
That is, the stochastic LTI system arising from the optimal predictor
is isomorphic to the stochastic LTI system from eq. \eqref{eq:assumedSys:innov}, Section \ref{sect:LTIaspred}.
Note that minimality and uniqueness 
of stochastic LTI systems for the case when 
there is feedback is much less understood, see \cite[Chapter 17]{LindquistBook} for
a detailed discussion on the difficulties arising in the presence of feedback. 
\\
Let now $\hat{f}=f_{(\hat{A},\hat{B},\hat{C},\hat{D})}$ be a predictor from $\mathcal{F}$
such that the generalization loss $\bE[\|\y(t)-\hat{\y}_{\hat{f}}(t)\|_2^2]$ is small. 
We do not require the absence of feedback from $\y$ to $\mathbf{u}$. 
Moreover, assume that  the stochastic system from eq. \eqref{eq:assumedSys:innov}
is such that  the pair $(C,A)$ is observable, the pairs
$(A, K)$ and $(A,B)$ are controllable. In addition, assume that $\w$ is coercive. Furthermore, assume that
$(\hat{C},\hat{A})$ is observable, and the pairs
$(\hat{A},\hat{B}_u)$ and $(\hat{A},\hat{K})$ are controllable,
where $\hat{B}=\begin{bmatrix} \hat{K},\hat{B}_u \end{bmatrix}$ and $\hat{B}_u$ has
$n_{\mathrm u}$ columns. 
In particular, the deterministic LTI system $(\hat{A},\hat{B},\hat{C},\hat{D})$ representing $f$
is minimal, and the deterministic LTI system $(\hat{A}_0,\hat{B}_0,\hat{C}_0,\hat{D}_0)$
representing the predictor associated with the stochastic LTI in eq. \eqref{eq:assumedSys:innov}, Section \ref{sect:LTIaspred}
is also minimal. 
Then the  inequalities \eqref{eq:pred:var1}-\eqref{eq:pred:var2} are also true for the case
 $\w=\begin{bmatrix} \y^T & \mathbf{u}^T \end{bmatrix}^T$, with the difference that
 $H_1=\hat{C}_0(zI-\hat{A}_0)^{-1}\hat{B}_0+\hat{D}_0$ and instead of
 $\y^d(t)$ one should use $\y(t)$.
 In particular, 
 \begin{equation}
 \label{eq:pred:var4}     
 \bE[\|\y(t)-\hat{\y}_f(t)\|^2_2] \ge \left(\|\hat{D}_0-\hat{D}\|_F^2+\sum_{k=0} \|\hat{C }_0\hat{A}^{k}_0\hat{B}_0-\hat{C}\hat{A}^{k}\hat{B}\|_F^2 \right) \mu_{w}
 \end{equation}
 where $\mu_w$ is a lower bound on the minimal eigenvalue of the spectral density of $\w$. Since $\w$ is coercive, then
 the spectral density of $\w$ is strictly positive definite, i.e., $\mu_w > 0$. 
 Hence,  if the generalization loss of $f$ 
 is small, then the $H_2$ distance
 between the deterministic LTI systems $(\hat{A}_0,\hat{B}_0,\hat{C}_0,\hat{D}_0)$ 
 and$(\hat{A},\hat{B},\hat{C},\hat{D})$ is also small. The former system
 describes the predictor associated with the stochastic LTI eq. \eqref{eq:assumedSys:innov}, and the latter
 deterministic LTI system  represents the predictor $\hat{f}$.
 Then from \cite[Theorem 3.2.1]{RalfPeeters}, using minimality of the deterministic LTI systems
 $(\hat{A},\hat{B},\hat{C},\hat{D})$ and $(\hat{A}_0,\hat{B}_0,\hat{C}_0,\hat{D}_0)$,
 it follows that if the generalization loss of $f$ is sufficiently small, then 
  for a suitable  nonsingular matrix $\mathcal{T}$, the norms of the differences
  $\|\hat{A}-\mathcal{T}^{-1}\hat{A}_0\mathcal{T}\|_F$,
  $\|\hat{B}-\mathcal{T}^{-1}\hat{B}_0\|_F$, $\|\hat{C}-\hat{C}_0\mathcal{T}\|_F$, $\|\hat{D}_0-\hat{D}\|_F$ are small. 
  Since in eq. \eqref{eq:assumedSys:innov} of Section \ref{sect:LTIaspred} the matrices $A,B,K,C$ can be replaced by
  $\mathcal{T}^{-1}A\mathcal{T},\mathcal{T}^{-1}B,\mathcal{T}^{-1}\mathcal{K},\mathcal{C}\mathcal{T}^{-1}$ without changing the output of eq. \eqref{eq:assumedSys:innov}, then, by taking into account
  the definition of $\hat{A}_0,\hat{B}_0,\hat{C}_0,\hat{D}_0$
  it follows that
  the norms $\|(\hat{A}_0+\hat{K}\hat{C})-A\|_F$, $\|(\hat{B}_u+\hat{K}\hat{D})-B\|_F$, $\|K-\hat{K}\|_F$, $\|C-\hat{C}\|_F$, $\|D-\hat{D}_u\|_F$ are small, where
  $\hat{D}=\begin{bmatrix} 0 & \hat{D}_u \end{bmatrix}$. 
  In other words, the matrix $\hat{A}_0+\hat{K}\hat{C}$ can be interpreted as an approximation
  of $A$, the matrix $\hat{B}_u+\hat{K}\hat{D}$ as an approximation of $B$,
  the matrix $\hat{K}$ as an approximation of $K$, the matrix $\hat{C}$ as an approximation of $C$ and the matrix $\hat{D}_u$ as an approximation  of $D$.
 
  To sum up, we have argued that if a predictor $f$ has a small generalization
  loss, then the matrices of the deterministic LTI system which represent $f$ can be
  used to compute an approximations of the matrices of a stochastic LTI system describing $\y$.
  Our results on PAC-Bayesian-Like bounds indicate that if $f$ has a small empirical 
  loss, then the generalization error will also be small. Hence, if $f$ has a small 
  empirical loss, then the matrices of the LTI system representing $f$ can be used to
  obtain approximations of a stochastic LTI system representing $\y$.

\end{document}